\documentclass[letterpaper, 10 pt, conference]{ieeeconf}  

\IEEEoverridecommandlockouts                              
\usepackage{svg}
\usepackage{amsmath}
\usepackage{amsfonts}
\usepackage{booktabs,siunitx}

\usepackage{graphicx}
\usepackage{comment}
\usepackage[caption=false]{subfig}
\let\labelindent\relax
\usepackage{enumitem}
\usepackage{todonotes}
\usepackage{url}

\usetikzlibrary{shapes, arrows.meta, positioning}

\title{\LARGE \bf
Impact of Object Weight in Handovers: Inspiring Robotic Grip Release and Motion from Human Handovers}

\author{Parag Khanna$^{1}$, Mårten Björkman$^{1}$ and Christian Smith$^{1}$
\thanks{$^{1}$ Division of Robotics, Perception and Learning (RPL), EECS, KTH Royal Institute of Technology, Sweden
        {\tt\small paragk@kth.se,} {\tt\small celle@kth.se,} {\tt\small ccs@kth.se}}%
}

\usepackage{hyphenat}
\begin{document}

\maketitle
\thispagestyle{empty}
\pagestyle{empty}

\begin{abstract}
This work explores the effect of object weight on human motion and grip release during handovers to enhance the naturalness, safety, and efficiency of robot-human interactions. We introduce adaptive robotic strategies based on the analysis of human handover behavior with varying object weights. The key contributions of this work includes the development of an adaptive grip-release strategy for robots, a detailed analysis of how object weight influences human motion to guide robotic motion adaptations, and the creation of handover-datasets incorporating various object weights, including the YCB handover dataset. 
By aligning robotic grip release and motion with human behavior, this work aims to improve robot-human handovers for different weighted objects. We also evaluate these human-inspired adaptive robotic strategies in robot-to-human handovers to assess their effectiveness and performance and demonstrate that they outperform the baseline approaches in terms of naturalness, efficiency, and user perception.  
\end{abstract}

\section{Introduction}
Given rapid technological advancements, collaborative robots are becoming increasingly present in daily human social environments, including manufacturing floors, retail stores, and homes \cite{Goodrich2008,survey_review_2022_object_handovers}.
As these robots become more integrated into our lives, they must master basic human interactions to be effective collaborators.
One such crucial interaction is the handover, a regular and necessary component of daily human activities that robots must perfect to collaborate smoothly with humans \cite{When_where_how_human-human_studyStrabala}.

Handovers consist of two primary phases: the reaching phase, where the giver's and the receiver's hands reach toward a common handover location, followed by the physical interaction as the receiver forms a grip on the object and the giver releases the object, referred to as ``grip release" \cite{survey_review_2022_object_handovers,Huber2008}.
Throughout their lives, humans acquire and enhance handover skills through countless experiences, effortlessly adjusting to objects of various weights during handovers \cite{Mason2005}. 
To achieve similar proficiency in handovers, robots must be capable of adapting their handover motions and grip-release mechanisms to objects of varying weights \cite{survey_review_2022_object_handovers,Parastegari2017,dataset-khanna}.
In this work, we focus on adapting robotic motion and grip release with respect to object weight, aiming to enhance the efficiency and naturalness of human-robot handovers.

\begin{figure}[t]
      \centering
     \includegraphics[width=\columnwidth,trim={0.0cm 1.5cm 0.0cm 3.50cm},clip]{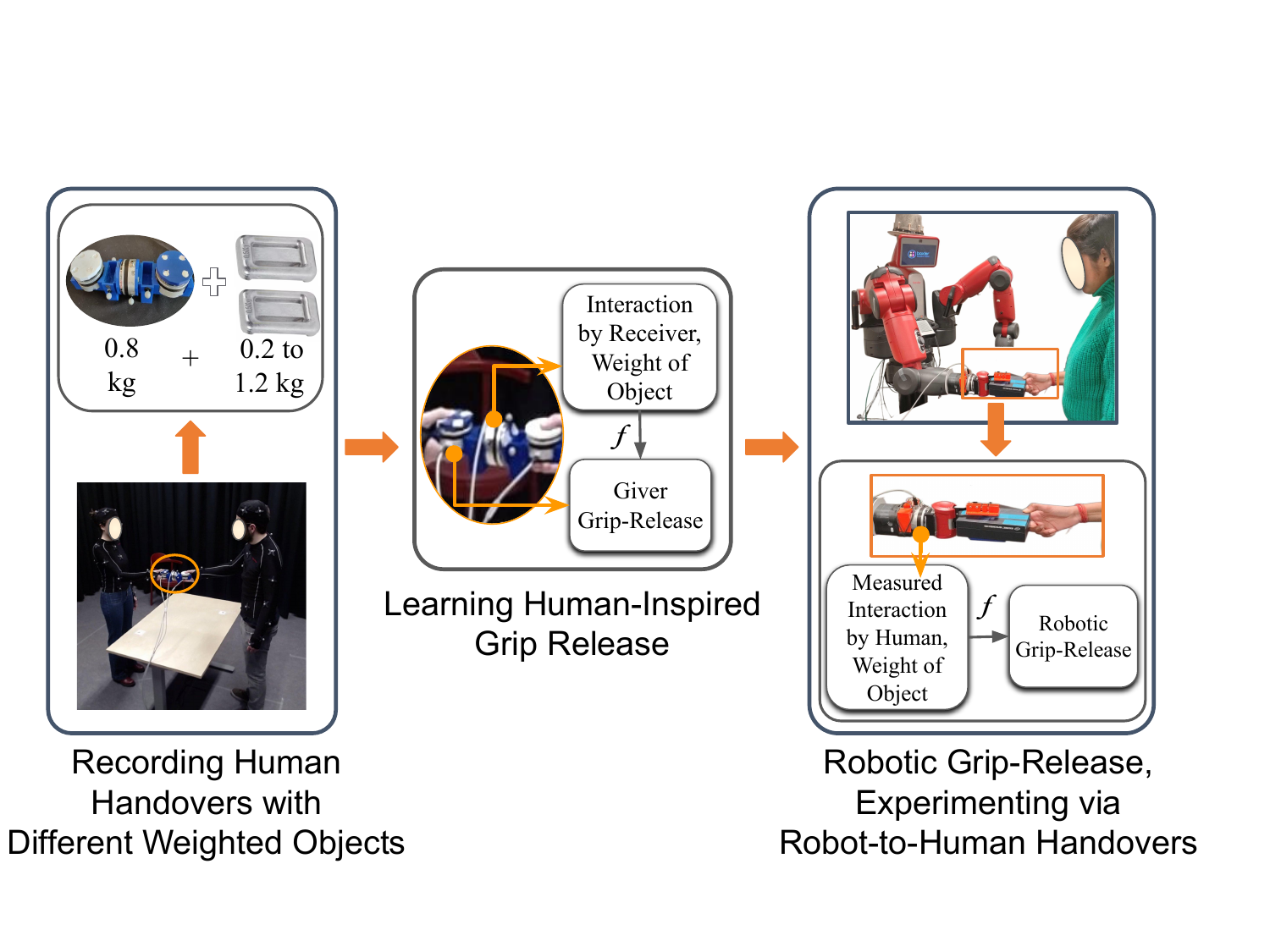}  
       \setlength\abovecaptionskip{-0.4\baselineskip}
      \caption{An overview of our approach: We investigate the impact of object weight on human motion and grip-release dynamics in human-human handovers. Based on these insights, we propose a data-driven strategy for adaptive grip release and evaluate its effectiveness in robot-human handovers. Additionally, we develop and assess weight-adaptive motion strategies for robots during handovers. This comprehensive approach aims to enhance the naturalness and efficiency of robot-human object handovers across various weight categories.
      }
      \label{fig:H2H_handover_study}
   \end{figure}

Due to their proficiency and adaptability, humans make ideal subjects for studying handovers. Numerous studies \cite{chan_grip_from_load_second_PR2,Modelling_human_reaching_phase_inH2H_forR2H_sina,When_where_how_human-human_studyStrabala} suggest that humans prefer robot behavior that mimics human actions. Therefore, this work focuses on learning from human handovers to inspire natural, safe, and efficient handovers between humans and robots. We present experimental studies of human handovers to create datasets and analyze the effect of object weight on human motion and grip-release during handovers and propose adaptive robotic grip-release and motion for handovers, as shown in Fig. \ref{fig:H2H_handover_study}.

In our previous works, we collected grip forces, interaction forces, and motion data during human-human handovers using a sensor-embedded object (0.8 kg) to create a multimodal dataset \textbf{Handovers@RPL} \cite{dataset-khanna}. Analyzing interaction forces preceding the grip-release by a human giver, we developed a human-inspired grip-release strategy \cite{parag-humanoids}. This strategy, used in robot-to-human handovers, commands the robot's grip-release based on interaction forces measured by a wrist sensor on the robot's arm. While this strategy outperformed baselines for the dataset (sensor-embedded) objects, it struggled to generalize across other objects of varying weights. In this work, we expand our multimodal dataset to include more weight classes of transferred objects (1.0, 1.2, 1.4, 1.6, and 2.0 kg), refeered to as \textbf{Handovers@RPL-2.0}, capturing the effect of weight on grip forces and human motion during handovers. Using the new data, particularly the forces observed, we formulate a grip-release strategy that takes into consideration the weight of the object in handover and appropriately adapts the robotic grip-release. 

The aforementioned dataset is with a specific sensor-embedded object with varying weight. To better generalize our analysis across different objects, we introduce the \textbf{YCB-Handovers} dataset, which documents human handovers involving various everyday objects from the YCB object dataset, weighing from 8 g to 2060 g. This dataset primarily aims to study the effect of varying object weights on human motion during handovers. We aim to answer two key questions: ``Can a robot observe human motion to accurately infer the weight or changes in weight of an object during a human-to-robot handover?" and ``How can robotic motion be adapted to effectively convey changes in object weight during a robot-to-human handover?". Based on our analysis, we propose corresponding changes to robotic motion to enhance the conveyance of object weight changes during handovers.

Thus, our contributions are:
\begin{enumerate}
    \item An adaptive robotic grip-release strategy tailored to object weight in robot-human handovers. 
    \item A comprehensive analysis of the impact of object weight on human motion, leading to human-inspired adaptive robotic motion. 
    \item Development of human handover datasets featuring objects of varying weights, including a dataset with grip forces-interaction forces and the YCB handover dataset.
\end{enumerate}

\section{Background and Related Work}
A handover task comprises multiple phases that humans perform seamlessly through collaboration. Both verbal and nonverbal communication are used to initiate the coordinated spatio-temporal movement of a giver and a taker. In a general handover, the giver determines a suitable location to transfer the object within the shared interpersonal space and starts moving the object towards that location \cite{human_human_to_human_robot_study_Controzzi} \cite{survey_review_2022_object_handovers}. The taker prejudges the handover location based on the giver's motion and approaches the location with their preferred hand. As the handover progresses, both the giver and taker adjust their arm speed based on visual sensory information. The coordinated movement ends with a light impact on the object as the taker begins to interact with it, forming a grip. During this interaction, the giver reduces its grip force on the object while the taker's grip force increases. Both parties share the responsibility of supporting the object's weight and maintaining its pose to prevent it from falling. The handover concludes when the giver releases the object completely after ensuring the taker has a stable grasp. Therefore, a robot must be competent in multiple phases of handover to achieve fluent exchanges.

Humans adapt swiftly to complex handover scenarios and different-weighted objects encountered in daily life, such as handing over a knife or a hammer with ease \cite{survey_review_2022_object_handovers, When_where_how_human-human_studyStrabala, object_orientation_dataset_chan}. There are scenarios where the weight of the object to be transferred differs from expectations, such as an unexpectedly heavy box. Human givers and takers quickly adapt to these uncertainties in weight. In certain cases, humans rely solely on haptic feedback to perform the handover \cite{dataset-khanna}. Even in these scenarios, both giver and taker adapt appropriately to ensure a safe and efficient handover. A social robot is likely to encounter such dynamic social environments and must adapt to ensure safe handovers while avoiding failures.

Interest in human-inspired handovers for robots has led to various studies on human handovers \cite{survey_review_2022_object_handovers}. These studies have explored human motion modeling, grip forces, and task-specific dynamics to inform robot designs.
\subsection{Datasets}
Several datasets support research in human-robot handovers. Carfi et al. \cite{dataset_Emaro_CARFI2019109} provided a dataset with 3D upper skeleton tracking using MoCap and smartwatch inertial data. The object orientation dataset by Chan \cite{object_orientation_dataset_chan} offers MoCap data of upper skeletons and object poses, aiding studies on proper object orientation during robot-to-human handovers. Precision grasp preferences were highlighted in a dataset by Cini et al. \cite{more_precision_type_grasp_type_location_study_dataset_cini_controzzi}. The H2O dataset presented by Ye et al. consists of annotated videos of handovers between 15 individuals for over 30 different objects, supporting vision-based tasks and providing insight into human-human handover dynamics. However, it lacked insights based on weight diversity. Other datasets, such as OHO \cite{stephan2023oho} and the Human-Object-Human (HOH) dataset \cite{wiederhold2023hoh}, provide multimodal data for handover analysis. The HOH dataset, encompassing 2,720 handovers involving 136 objects and 40 participants, offers comprehensive multi-modal data to analyze handover mechanics but does not explicitly include object weights. HandoverSim, a Python-based simulator introduced by Chao et al. \cite{chao2022handoversim}, enables controlled handover studies but does not fully capture real-world complexities. 
In our prior work \cite{dataset-khanna}, we recorded a first-of-its-kind multimodal dataset of human handovers, \textit{Handovers@RPL},  capturing both human motion and the forces involved in the transfer process. Thirteen pairs of participants passed an object embedded (0.8 kg) with force/torque (F/T) sensors in a motion-capture environment, allowing us to measure grip forces for both the giver and taker, as well as interaction forces. We specifically investigated the effects of increasing the object's weight by 1 kg (1.8 kg), observing significant differences in key handover properties such as the time of transfer and time of grip-release.

\begin{table*}[t]
\centering
\caption{Summary of Handovers and YCB Handovers: Details of Objects in Different Baskets}
\begin{tabular}{|l|c|c|c|c|c|}
\hline
\multicolumn{6}{|c|}{\textbf{Summary of Handovers with Sensor-Embedded Baton}} \\ \hline
\textbf{Handover Type and Weight} & \textbf{No. of Handovers} & \textbf{Forces} & \textbf{Motion} & \textbf{Weight (kg)} & \textbf{ Object Details} \\ \hline
\textbf{New Data Handovers} & 3235 total & Yes & Yes & - & - \\ \hline
\hspace{0.5cm}  & 671 & Yes & Yes & 1.0 & New Data - Baton + 0.2 kg\\ 
\hspace{0.5cm}  & 654 & Yes & Yes & 1.2 & New Data - Baton + 0.4 kg \\ 
\hspace{0.5cm}  & 662 & Yes & Yes & 1.4 & New Data - Baton + 0.6 kg \\ 
\hspace{0.5cm}  & 601 & Yes & Yes & 1.6 & New Data - Baton + 0.8 kg \\ 
\hspace{0.5cm}  & 647 & Yes & Yes & 2.0 & New Data - Baton + 1.2 kg \\ \hline
\textbf{Previous Data Handovers} & 5763 total & Yes & Yes & - & - \\ \hline
\hspace{0.5cm}  & 2999 & Yes & Yes & 0.8 & Previous Data - Baton\\ 
\hspace{0.5cm}  & 2764 & Yes & Yes & 1.8 & Previous Data - Baton + 1.0 kg\\ \hline
\multicolumn{6}{|c|}{\textbf{YCB Handovers: Details of Objects in Different Baskets}} \\ \hline
\textbf{Basket} & \textbf{No. of Handovers} & \textbf{Forces} & \textbf{Motion} & \textbf{Weight (kg)} & \textbf{Object Details} \\ \hline
1 & 128 & No & Yes & 0.008 & Marker Small (MS) \\ 
  & 128  & No & Yes & 0.118 & Mug (M) \\ 
  & 128  & No & Yes & 0.242 & Wrench (W) \\ 
  & 126  & No & Yes & 0.600 & Mustard Bottle (MB) \\ 
  & 128  & No & Yes & 1.131 & Cleanser Bottle (CB) \\ \hline
2 & 108 & No & Yes & 0.040 & Racquetball (R) \\ 
  & 106    & No & Yes & 0.190 & Jello-Choc Box (J) \\ 
  & 106    & No & Yes & 0.374 & Meat Can (MC) \\ 
  & 107   & No & Yes & 0.874 & Hand Drill (HD) \\ 
  & 107    & No & Yes & 1.020 & Spray Bottle (SB) \\ 
  & 103    & No & Yes & 1.450 & Earphone Cover with Weights (E) \\ \hline
3 & 95 & No & Yes & 0.055 & Spatula (SP) \\ 
  & 101  & No & Yes & 0.149 & Bowl (B) \\ 
  & 99  & No & Yes & 0.202 & Clamp (C) \\ 
  & 101  & No & Yes & 0.410 & Coffee Can (CC) \\ 
  & 100  & No & Yes & 0.728 & Wood Block (WB) \\ 
  & 100  & No & Yes & 1.300 & Black Box with Weights (BW) \\ \hline
4 & 99 & No & Yes & 0.015 & Large Marker (LM) \\ 
  & 98 & No & Yes & 0.095 & Screwdriver (SC) \\ 
  & 100 & No & Yes & 0.183 & Pitcher Base Y (P) \\ 
  & 101 & No & Yes & 0.514 & Sugar Box (SBX) \\ 
  & 100 & No & Yes & 0.608 & Hammer (H) \\ 
  & 85 & No & Yes & 0.925 & Skillet (SK) \\ \hline
5 & 91 & No & Yes & 2.060 & Pitcher with Added Weights (Heavy Weight-Not Careful) \\ 
  & 69 & No & Yes & 2.060 & Pitcher with Water (Heavy Weight-Careful) \\ 
  & 96 & No & Yes & 0.008 & Measuring Cup (Light Weight-Not Careful) \\ 
  & 61 & No & Yes & 0.048 & Measuring Cup with Water (Light Weight-Careful) \\ \hline
\end{tabular}
\label{tab:handover_summary}
\end{table*}

\subsection{Human Motion and Joint Coordination}
Several studies model human motion to inspire robotic arm movements.
To better understand verbal and non-verbal cues, as well as joint coordination, Strabala et al. \cite{When_where_how_human-human_studyStrabala} utilized multiple color and depth cameras to study human-human handovers. Rasch et al. \cite{human-like-motion-forR2H_Handovers_Rasch2018AJM} recorded human arm motions to develop a joint motion model for robotic givers. Another study employed an electromagnetic tracker and markers to evaluate human hand motions and inspire robot reaching profiles \cite{robot_reaching_profiles_4600651}. Aleotti et al. \cite{aleotti2014affordance} optimized object orientation during handovers. Chang \cite{chang2021learning} introduced a deep reinforcement learning approach for predicting human-preferred grasps, and Lori et al. \cite{iori2023dmp,perovic2023adaptive} used Dynamic Movement Primitives and Preference Learning for adaptive handover control. 
The effect of object weight on handover location and duration was investigated in \cite{h2H_handovers_how_dis_and_obj_mass_matter_Clint2017}, showing that only handover-duration was impacted. 

Studies have shown that humans adapt their movements when handling objects of different weights or contents. In human-robot handovers, enabling robots to infer an object's weight by observing human motion is essential for improving robotic preparedness.
It was demonstrated that humans can estimate object weight by observing both human and humanoid lifting actions, highlighting the potential for robots to assess weight through visual cues \cite{Sciutti2014_weight_motion}. 
An interesting finding in \cite{careful_handovers_Lastrico} indicated that humans exhibit careful motion in handovers, even during the reach-to-grasp phase when handling cups filled with water as opposed to empty cups. This carefulness can be reliably detected online before the handover action is completed. It was further proposed that robot motion could be modulated to express awareness of the object's properties, such as whether a cup is full or empty, thereby conveying important information to human partners in advance and potentially improving task efficiency in human-robot interaction scenarios.

Furthermore, \cite{humanoids_simulated_motion_objectweight2009} investigated how human-inspired handover movements can convey object weight in simulated exchanges between humanoid robots. Bimanual human handover movements involving objects of varying weights (2, 5, 10, and 15 kg) were analyzed and replicated in a simulation of two human-like robots. In a user study, participants watched the simulation videos and answered a questionnaire assessing their perception of object weight. Results showed that while participants could distinguish between heavy and light objects, estimating the exact weight remained challenging.

\subsection{Forces in Handovers and Grip Release}
For studying forces in handovers, \cite{chan_grip_from_load_second_PR2} used a baton with grip and load force sensors, developing a controller for robotic grip forces based on the measured relationship between grip and load forces in vertical transfer of the baton. Controzzi et al. \cite{human_human_to_human_robot_study_Controzzi} observed that taker arm speed influences the giver's grip forces.
Studies show that human grip-release actions typically occur within 500 milliseconds \cite{chan_grip_from_load_baton}, \cite{study_in_person_handover_for_baton}, \cite{human_human_to_human_robot_study_Controzzi}, necessitating rapid decision-making for robots. While vision-based systems detect human hand approach and grip, they are challenged by occlusions and the variability in human grasps \cite{many_grasp_types_7243327}, \cite{grasp_classification_vision_DBLP:conf/iros/YangPCF20}. Capacitive proximity sensors \cite{capacitive_sensors_grasp_detect_9560970} can overcome some of these limitations by detecting human grasps without visual input. In our prior work \cite{dataset-khanna}, we studied the effect of object weight on the grip release and the interaction forces in handovers. We find that a 1 kg increase in the weight of object caused a significant difference in the grip-release time, the transfer time and the interaction forces (Pull force).

The most common strategy for commanding grip release for a robotic giver in robot-to-human handovers is based on \textit{Pull-Force}.
Vertical handovers often rely on grip-release triggered by a slight upward pull force detected by wrist sensors \cite{chan_grip_from_load_baton}, \cite{chan_grip_from_load_second_PR2}, \cite{chan_grip_from_load_humanoid-6907004}. For horizontal handovers as well, grip-release is typically based on thresholding the measured external forces on the robot wrist \cite{human_human_to_human_robot_study_Controzzi}, with pull force, measured along the horizontal transfer direction, used commonly \cite{survey_review_2022_object_handovers}. In another study \cite{pull_proactive_strategy_8673085}, proactive grip-release strategies based on grip force disturbances during a human grasp and pull-force deliver the best results. 
Another grip-release strategy involves \textit{Load-Sharing}. As humans hand over an object, its weight decreases due to load-sharing between the giver and taker, culminating in a full transfer. In \cite{loadsharing_pull_strategy-10.3389/frobt.2021.672995, loadshare_strategy-7803296}, the robot giver gets ready for grip release when the taker's shared load reaches 50\%. However, the actual grip release is triggered by detecting a pull force threshold.

A human-inspired, data-driven grip release strategy was proposed in our prior work, which relied on interaction forces (including pull force and load share) measured during handovers. This innovative approach was developed to enhance the naturalness and efficiency of robot-to-human object transfers. Based on a comprehensive dataset of interaction and grip forces observed in human-human handovers, this strategy utilized a Long Short-Term Memory (LSTM) network, a  recurrent neural network known for its ability to process and predict based on time-series data.
The LSTM was trained on time series of interaction forces observed preceding the grip release by human givers, allowing it to learn the subtle cues and patterns that humans use to coordinate the grip-release and taking actions during handovers. In robot-to-human handovers, as a human taker begins to take the object, this strategy utilizes the trained LSTM to process real-time interaction forces measured by a wrist-mounted force/torque sensor. By analyzing these forces, the system can predict the appropriate moment for grip release, commanding the robotic giver to let go of the object in a manner that closely mimics human behavior. This approach not only improves the timing and smoothness of the handover but also enhances the overall human-robot interaction experience by making it feel more natural and intuitive to the human recipient, which was shown in robot-to-human handover experimentation. However, this grip release strategy demonstrated limited generalizability across objects with weights differing from the training object used in the study. For objects significantly lighter or heavier than the training object,
it neither proved to be the fastest approach nor was it perceived as the most natural by participants during the evaluation process. This highlighted the need for a more sophisticated grip release strategy that adapts to the weight of the object.

The literature review above highlights a significant gap in the analysis of how object weight impacts human motion across a wide range of weights. There is limited research on the effects of weight on interaction forces and grip release during handovers across a wide range of weights as well. Additionally, further investigation is needed into how humans adapt their grip release strategies in response to varying object weights, as well as how robots can be inspired to implement similar adaptations. In our work, we aim to address these shortcomings by analyzing the influence of object weight on human motion and grip release, while also developing data-driven strategies for adaptive grip release in robots.

\section{Human Handovers Study and Datasets}
With the aim to analyse the forces and the motion in human handovers, we ran experimental studies where participants handed over different objects to each other in a motion capture room.

\subsection{Experimental Setup}
To record participant movements, the experiment was conducted in an Optitrack MoCap room \cite{optitrack+Doe:2022:Online}, with participants wearing upper body MoCap suits to track their movements. Standing across a table (Fig. \ref{fig:table_setup_and_precision_grasp}(a)), participant 1, as the giver, picked up an object from position A and handed it mid-air horizontally to participant 2, the receiver, who placed it at position B. Then, participant 2, as the giver, handed the object back to participant 1, who returned it to position A, repeating the procedure. To maintain naturalness, no instructions on speed or location of handover were given. Positions A and B were marked close to the participants' dominant hands to reduce ergonomic load, shown for two right-handed participants in Fig. \ref{fig:table_setup_and_precision_grasp}(a).

\subsection{Handovers with sensor-embedded object}
\subsubsection{Sensor-Embedded Object}
\begin{figure}[b]
  \centering
  \subfloat[]{
  \includegraphics[width=4.9cm,height=4cm,trim={5cm 4.4cm 4.5cm 3.3cm},clip]{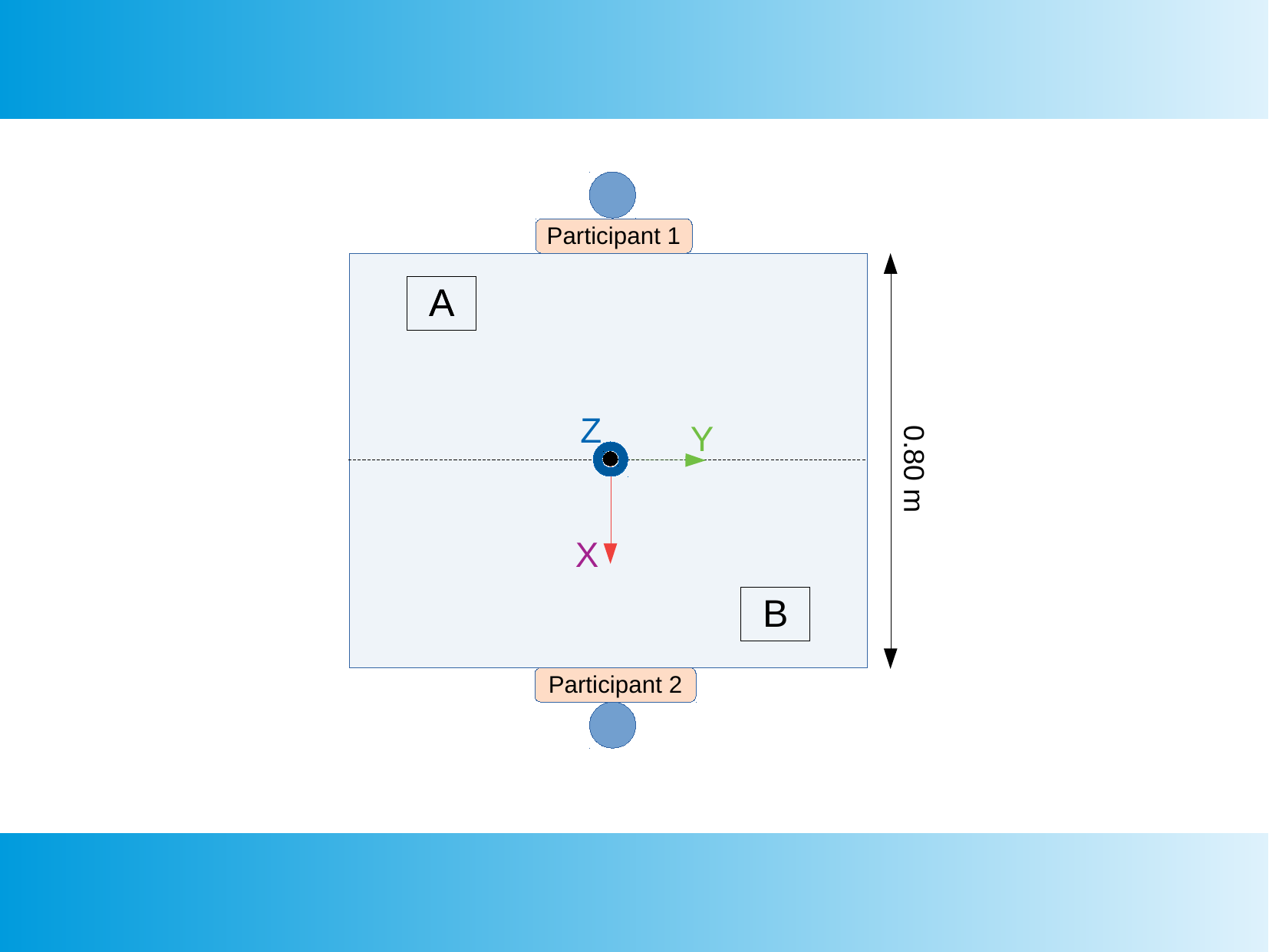}}
  \subfloat[]{
   \includegraphics[width=3.4cm,height=3.0cm,trim={4cm 4cm 4.5cm 8.0cm},clip]{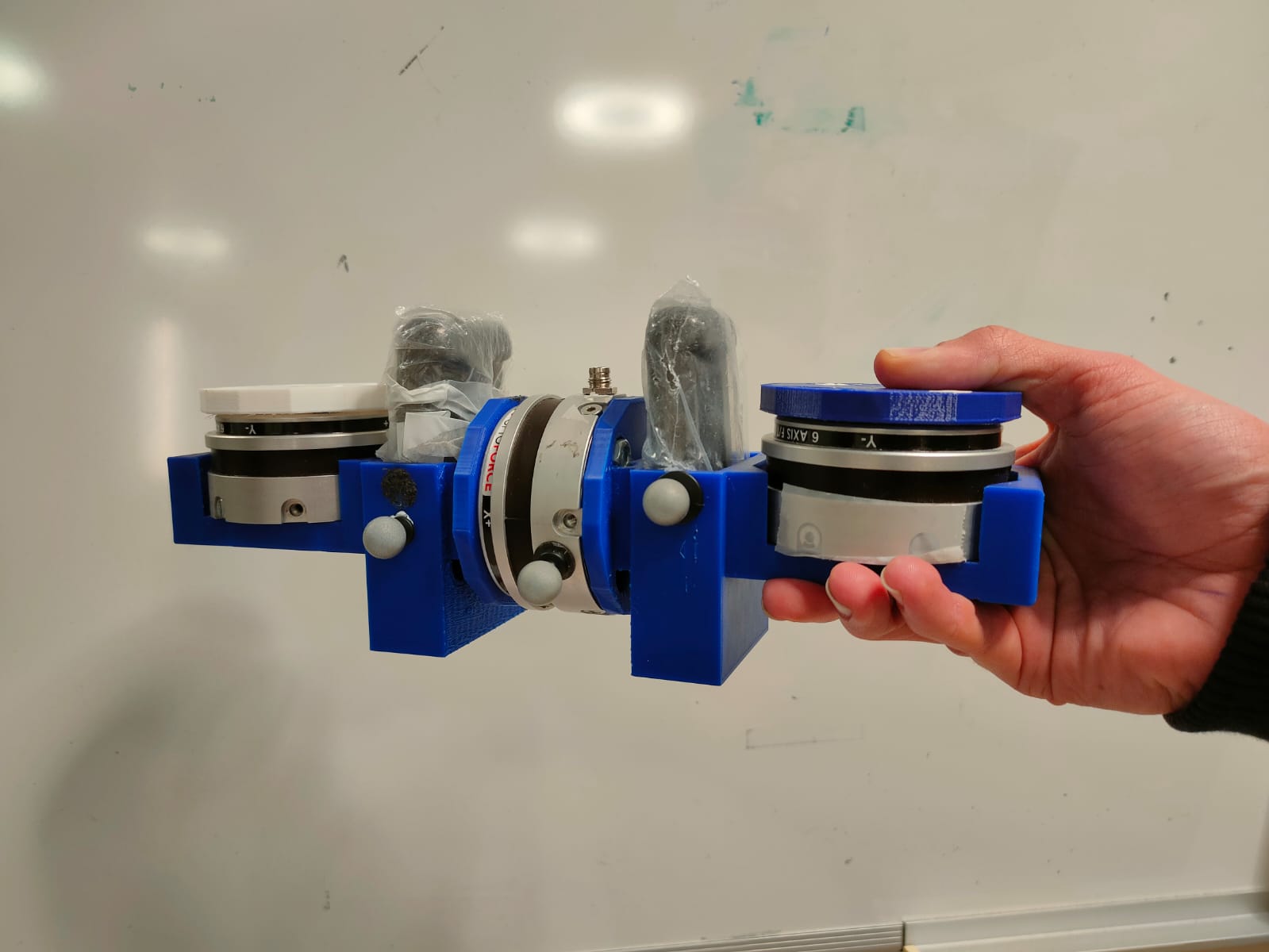}}
  \caption{ (a) Table (0.8 m wide) setup for human-human handovers (b) Precision grasp on baton with lead weights added}
  \label{fig:table_setup_and_precision_grasp}
\end{figure}

The experiment involved passing a 3D-printed, sensor-embedded baton between participants. The baton housed three 6D Force/Torque (F/T) sensors from Onrobot \cite{Onrobots_F_T+Doe:2022:Online}, each measuring a 6D wrench (3 forces and 3 torques). The central sensor measured interaction forces, while the side sensors measured grip forces. Five asymmetrically placed MoCap markers tracked the baton's movement. Weighing 0.8 kg and measuring 0.28x0.085x0.075 m, the baton also had slots for adding lead weights. The grip force sensors were color-coded blue and white for participant identification: participant 1 used the blue side and participant 2 used the white side, while the participants were instructed to use precision grasp as shown in Fig. \ref{fig:table_setup_and_precision_grasp}.

\begin{figure}[t]

  \centering
  \setlength\abovecaptionskip{-0.25\baselineskip}
  \includegraphics[scale=0.35,trim={4cm 1.8cm 4.5cm 4.1cm},clip]{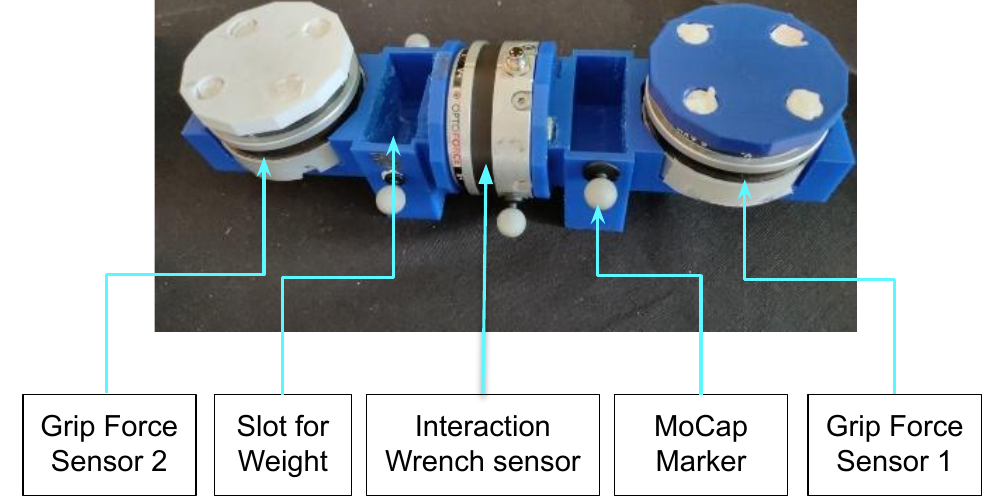}
  \caption{3D printed sensor embedded Baton}
  \label{fig:baton_details}
\end{figure}
The previously recorded dataset, \textit{Handovers@RPL} \cite{dataset-khanna} had 2 weight classes for the object in handovers: 0.8 kg and 1.8 kg, with data collected from 13 pairs. In the new study, we decided to introduce 5 more classes: 1.0, 1.2, 1.4, 1.6 and 2.0 kg by adding appropriate weights in the slot of the baton, to record a new dataset, \textbf{Handovers@RPL-2.0.}

\subsubsection{Experimental Settings}
For each participant pair, the experiment was conducted in five rounds, each corresponding to a different weight class. Participants were required to perform handovers for 9 minutes per round. Importantly, they had no prior knowledge of the baton's weight, and this information was not disclosed to them during the experiment.

\subsubsection{Participants}
Participants were recruited via advertisements on the university campus. The study involved 10 participants (5 pairs), all right-handed. Each participant received a 100 SEK gift voucher for their participation.
For all participants mentioned in this work, as per the local regulations, we are exempt from ethical approval as we did not collect any sensitive personal data (racial/ethnic origin, political views, religious/philosophical beliefs, health/sexual life) and this research doesn't involve physical intervention on or biological samples from participants.
In the absence of a relevant ethics board, we followed guidelines of the Declaration of Helsinki.
Participants began by completing a consent form for data collection; and reading the study instructions. 
Particularly, they consented to the use and distribution of their anonymized data and the use of collected video data 
in academic articles and presentations. 



\subsection{Handovers with generic Objects: YCB Handovers Dataset}
Since we aimed to capture handover forces in the aforementioned human-human handover datasets, we used a specially designed sensor-embedded object. This meant that all handovers were performed using a specific object with a fixed shape and size.

However, for a more comprehensive analysis of the impact of object weight in human handovers, it was essential to include objects of varying shapes and sizes, as well as those with more generic use in daily life. This would allow for a broader understanding of how object weight influences handover dynamics.

\subsubsection{YCB dataset objects}
The YCB Object and Model Set (Yale-CMU-Berkeley) is a widely-used benchmarking tool for robotic manipulation research consisting of different objects from everyday lives.
To have generic objects with varying weights, we selected various objects, primarily from the YCB dataset, as described in Table \ref{tab:handover_summary}.
Using objects from the YCB dataset for human handovers is beneficial because it provides a standardized and widely recognized set of objects that are commonly used in robotic manipulation studies. This \textbf{YCB-Handovers} dataset would allow for better comparison and generalization of our results across different research studies.
\subsubsection{Experimental Settings}
The objects were divided into five baskets, each containing items within a specific weight range. The experiment was conducted in five rounds, with each round corresponding to one basket. During a round, a basket filled with objects was placed at position A (as shown in the figure), and an empty basket at position B. Participant 1 handed over objects one by one from the basket at position A, while Participant 2 placed them in the empty basket at position B. Once Participant 1 finished handing over all objects from basket A, Participant 2 began handing over objects from basket B. This process was repeated for 9 minutes for each basket. 
In the final round, i.e., the fifth basket, we introduced four special objects to capture both careful and non-careful human-human handovers, along with the effect of added weights.  
Previous research \cite{careful_handovers_Lastrico} demonstrated that carefulness in human handover motions can be detected online when transferring filled versus empty cups.  
To extend this analysis, we incorporated a measuring cup (8g and 48g with water) as well as the YCB pitcher, filled either with water or additional weights.  
These handovers can be used to examine how added weight influences carefulness in human handover motions, however this analysis has not been included in this work.

\subsubsection{Participants}
For these experiments, 12 participants were recruited through advertisements on the university campus, forming six pairs. Five pairs consisted of right-handed participants, while one pair was a mix of a left-handed and a right-handed participant.
They each were reimbursed with a 100 SEK gift voucher.
\begin{figure}[t]
     \centering
    \setlength\abovecaptionskip{-0.4\baselineskip}
    \includegraphics[width=150pt,height=4.2cm,trim={0.1cm 5.5cm 1.7cm 6cm},clip]{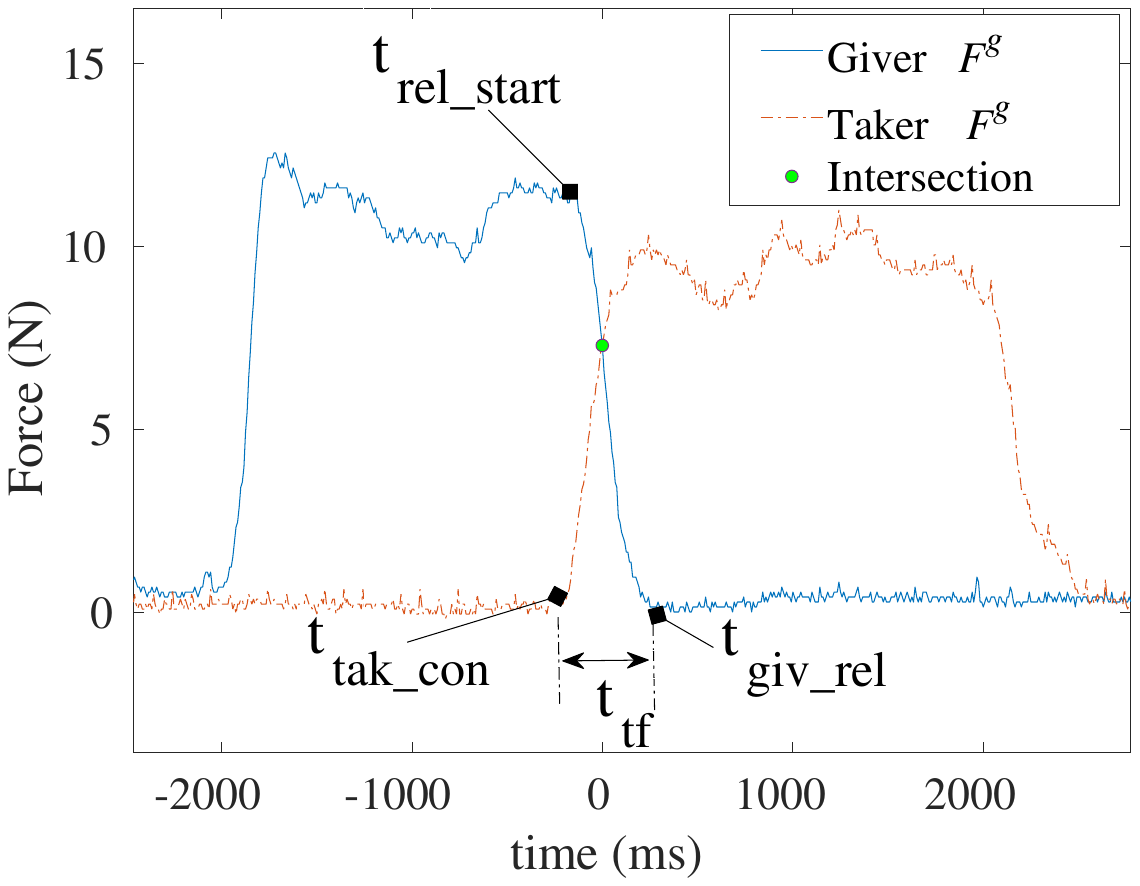}
      \caption{Grip force ($F^g$) variation in a particular handover.}
      \label{fig:forces_handover}
   \end{figure}

\subsection{Data Collection}
For each experiment, sensory data was recorded using the Robot Operating System (ROS) \cite{ROS-Doe:2022:Online}. 
MoCap tracked the baton's pose and participants' upper body skeletal segments (Hip, Ab, Chest, Neck, Head, Shoulders, Upper arms, Forearms, and Hands) at 120Hz. Data was broadcasted by Optitrack's MOTIVE software \cite{optitrack+Doe:2022:Online} and synchronized by a custom ROS node \cite{github_self_code}.
When using the sensor-embedded baton, F/T sensors provided 6D wrench data at 333Hz which was read using the aforementioned ROSnode, capturing interaction and grip forces. This resulted in synchronized data for F/T sensors, baton pose, and skeletal representation at 120Hz. 

\subsection{Datasets}
The continuous recordings of handovers for each pair were postprocessed to separate individual handovers. For each handover, as the giver releases the object and the taker takes it, the grip forces of the giver decrease to zero while the grip forces of the taker increase from zero, as shown in Fig. \ref{fig:forces_handover}. When the sensor-embedded baton (\textbf{Handovers@RPL-2.0}) was being used, we utilized this intersection of grip forces to segment each handover. For handovers with objects other than the sensor-embedded baton (\textbf{YCB-handovers}), we segmented the handovers when both giver and taker are holding the object during the handover, using motion tracking data. The separated handovers are saved for a duration of 6.666 seconds or 800 timesteps at 120 Hz. To recap, we collected and utilizeed the datasets of human handovers as described in Table \ref{tab:handover_summary}. These datasets are publicly available in a realted repository. \cite{parag_git_handover_datasets}.

\subsection{Features}
\begin{table}[h!]
\setlength\abovecaptionskip{-0.5\baselineskip}
  \caption{Signals of one recorded handover}
  \centering
  \label{tab:onehandover}
  \resizebox{0.8\columnwidth}{!}{
  \begin{tabular}{cccc}
    \toprule
    
    
    Signal & Signal & Signal  \\   
     No. & Name & Components \\
    
    \midrule
     1 & Wrench\_interaction & Force (x,y,z) \\
    & & Torque(x,y,z) \\
    2 & {Wrench\_giver}$^*$  & ---\texttt{"}--- \\
    3 & {Wrench\_taker}$^*$ & ---\texttt{"}---    \\
    4 & baton\_pose & Position(x,y,z)  \\
     & & {Rotation}($q_0,q_1,q_2,q_3$)\\
    5-17 & giver-Skeleton & 13 bodies-pose\\
    18-30 & taker-Skeleton & 13 bodies-pose\\
    \bottomrule
    \multicolumn{2}{c}{* Grip force is given by -Force(z) } 
    \end{tabular}
}
\end{table}
Table \ref{tab:onehandover} summarizes all information saved for an individual handover. It lists the signal names with which they are available in the data set. The wrench signals and the baton pose exist for the handovers with sensor-embedded baton only. The metadata includes the height, arm lengths, age, and handedness of the two subjects.
\section{Data analysis: Impact of Weight}

\subsection{Grip forces and Grip Release}
\begin{figure}[h]
  \centering
  \setlength\abovecaptionskip{-0.1\baselineskip}
  \includegraphics[width=0.8\linewidth,trim={3.5cm 8.1cm 4.69cm 9.0cm},clip]{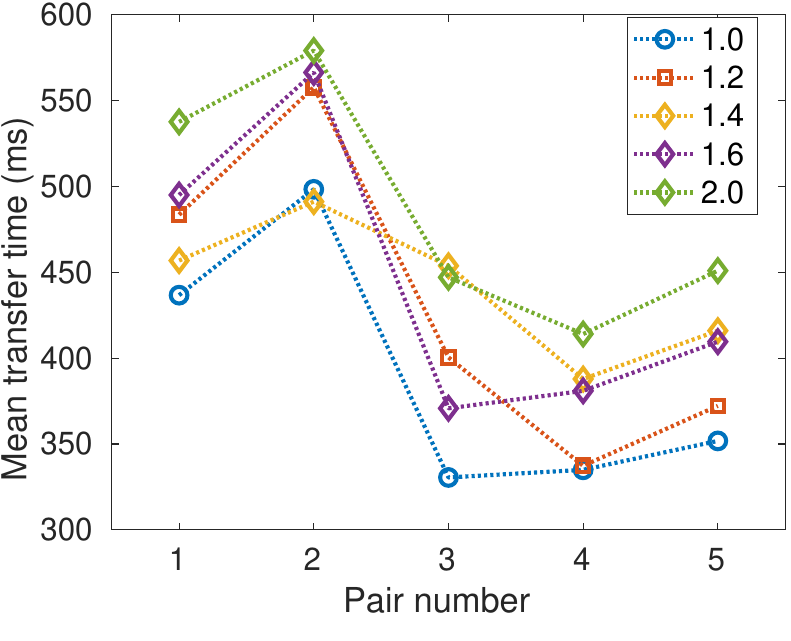}
  \includegraphics[width=0.8\linewidth,trim={0.7cm 6.8cm 2.3cm 7.3cm},clip]{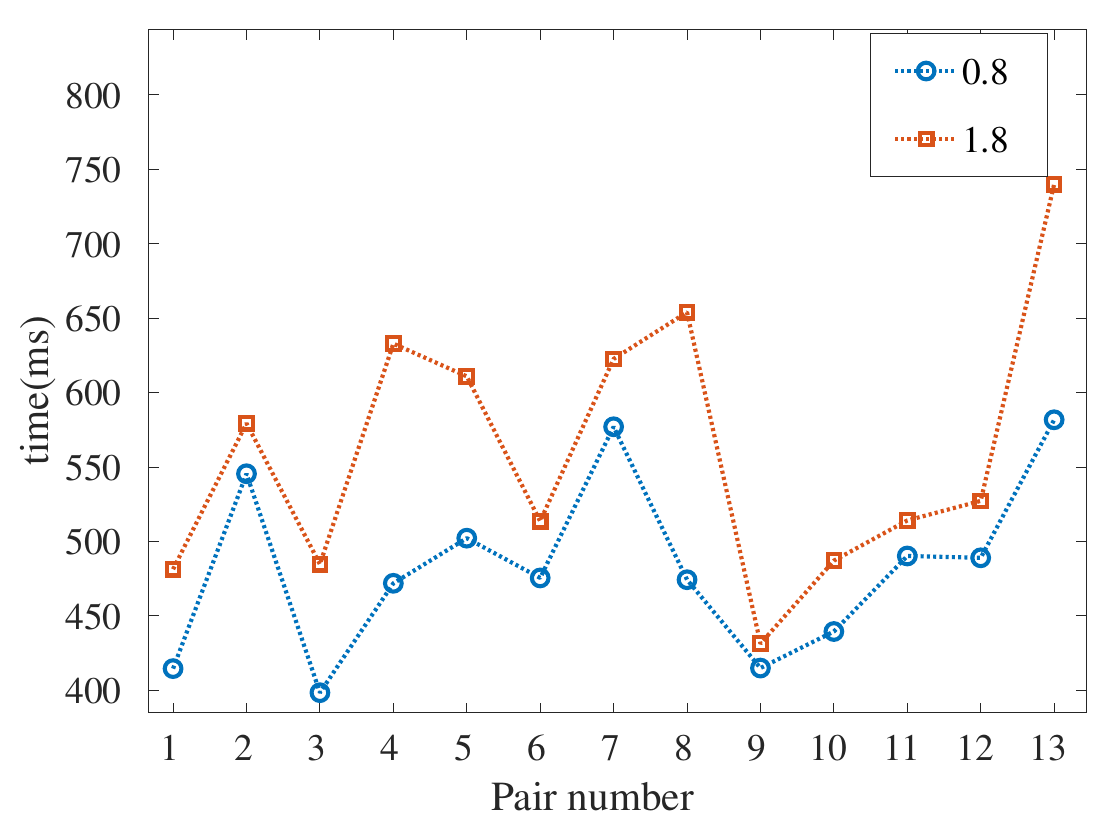}
  \caption{Mean transfer time}
  \label{fig:t_transfer}
\end{figure}
\subsubsection{Transfer time: $t_{tf}$}
The transfer time in a handover is defined as the interval from the taker's initial contact to the giver's final release \cite{chan_grip_from_load_second_PR2}. The taker contact time ($t_{tak\_con}$) is defined as when the taker's grip force ($F^g$) exceeds a 0.4 N threshold before $t = 0$ ms, and giver release time ($t_{giv\_rel}$) when the giver's grip force falls below this threshold. The transfer time ($t_{tf}$) is then calculated as: \begin{equation} t_{tf} = t_{giv\_rel} - t_{tak\_con} \nonumber \end{equation} Fig. \ref{fig:t_transfer} shows the mean transfer time variation for each pair across the different weighted objects. The increased object weight resulted in a longer mean $t_{tf}$. One-way ANOVA analysis (p$<$ 0.0001) confirmed significant differences between settings and the post-hoc t-test indicates significant increased $t_{tf}$ for all participants with a weight increase of 1 kg, for the new Handover@RPL-2.0 handovers as well as the previous dataset \cite{dataset-khanna}.

\subsubsection{Grip release time - $t_{gr}$}

\begin{figure}[h!]
  \centering
  \includegraphics[width=0.8\linewidth,trim={2.7cm 6.7cm 4.4cm 4.8cm},clip]{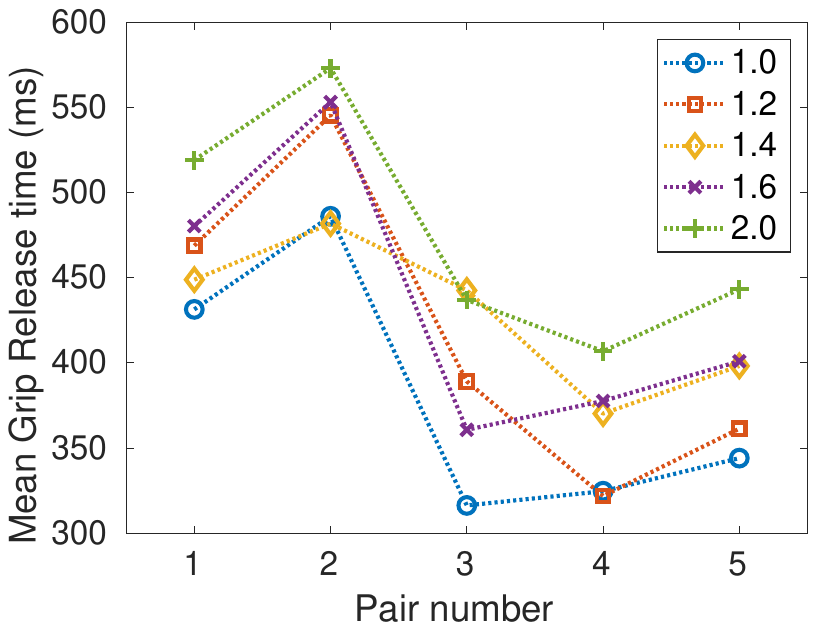}
  \vspace{-2.5mm}
  \includegraphics[width=0.8\linewidth,trim={0.7cm 6.7cm 2.3cm 5.3cm},clip]{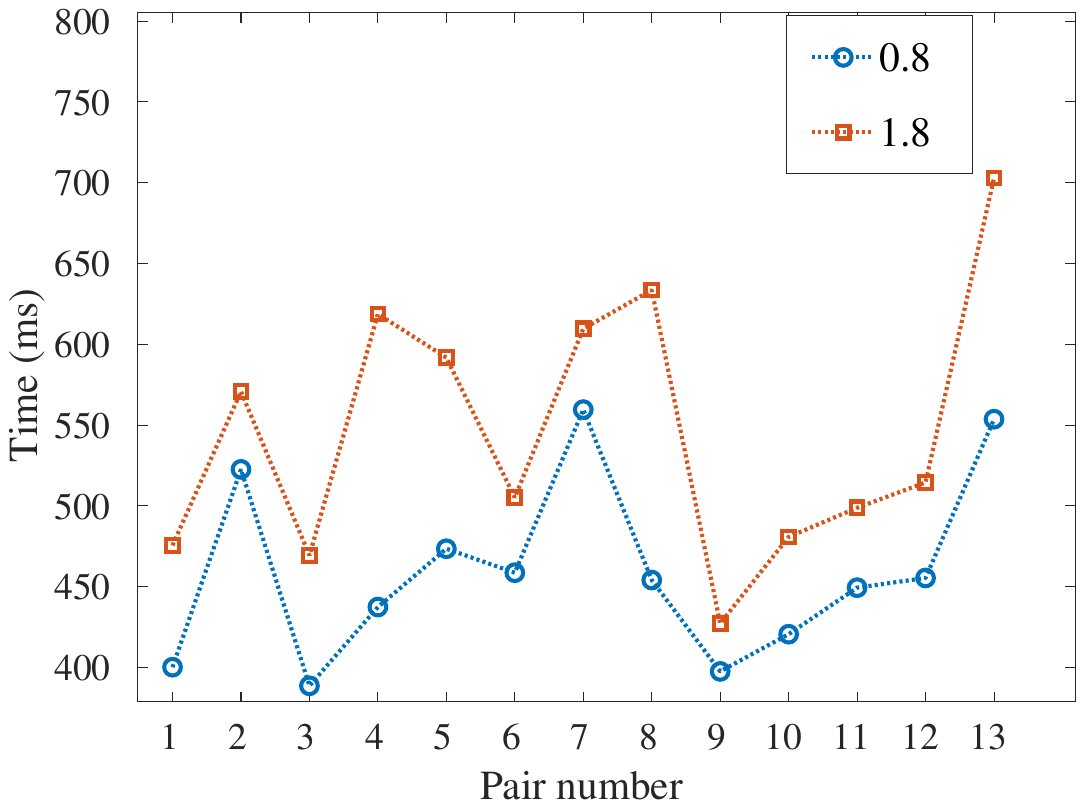}
  \caption{Mean grip release time}
  \label{fig:time_grip_release_new}
\end{figure}
Grip release time, the duration a giver takes to release the object during a handover, is crucial for designing robotic givers based on human data. This time is determined by measuring the period during which the giver's grip force ($F^g$) decreases to zero. The grip release begins with the first observed decrease in giver-$F^g$ following the taker's contact and ends when the giver's forces drop below the minimal threshold (at time $t_{giv\_rel}$ in Fig. \ref{fig:forces_handover}).

We observe that for most pairs, $t_{gr}$ increases with the weight of the object in Fig. \ref{fig:time_grip_release_new}, in line with the  analysis for the previous dataset \cite{dataset-khanna}. The results of a one-way ANOVA analysis with pair number as a random effect (p$<$0.001) show that the difference is found to be significant across all participants, and post hoc t-tests show a significant increase in $t_{gr}$ with a weight difference of 1 kg.

Therefore, we can conclude that the lightest object consistently results in the fastest grip release and transfer time, while the heaviest object leads to the slowest times across all pairs in both the datasets. The intermediate objects fall in between, showing a general trend of increasing times with increasing weight.

\subsubsection{Interaction Forces - Pull force}
\begin{figure}[h]

    \centering
    \subfloat[]{\includegraphics[width=.45\linewidth,height=3.5cm,trim={0.1cm 5.8cm 1.5cm 6cm},clip]{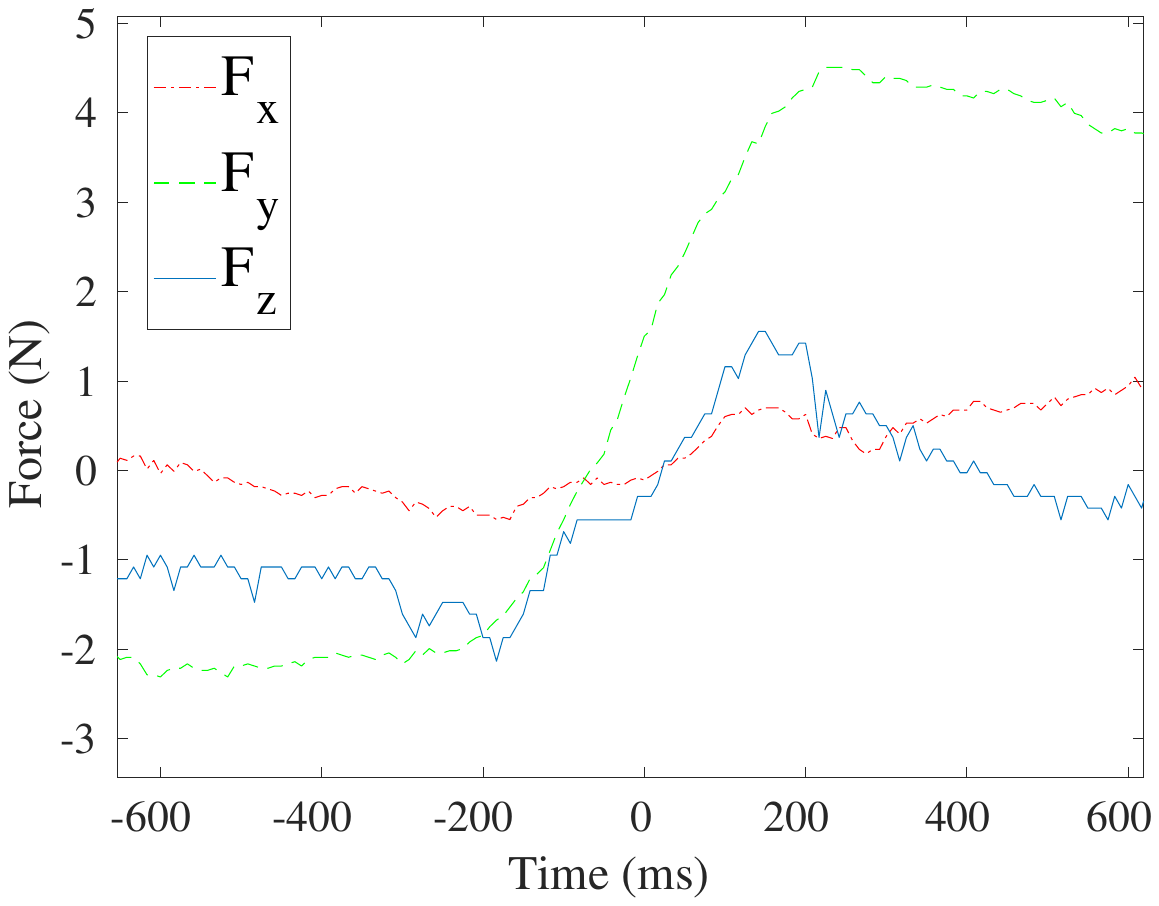}}
    \subfloat[]{\includegraphics[width=.54\linewidth,height=3.2cm,trim={{2.0cm 3.7cm 4.0cm 4.7cm}},clip]{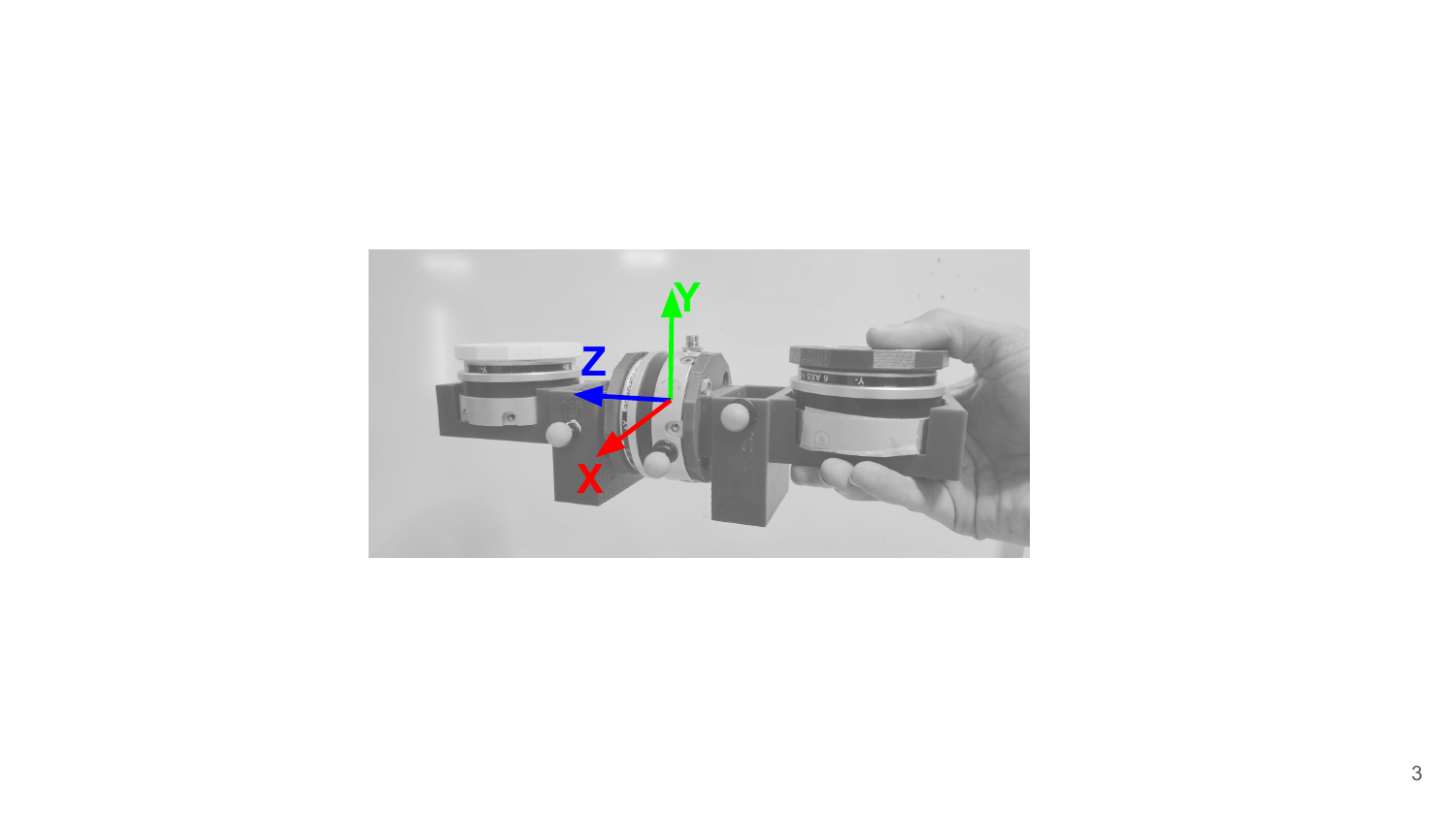}}
   
    \caption{Interaction Force: (a) Plot (b) Interaction sensor}
    \label{fig:intx_forces_handover}
\end{figure}

\begin{figure}[h]
  \centering
          \setlength\abovecaptionskip{-0.3\baselineskip}     
          \includegraphics[width=0.8\linewidth,trim={2.4cm 6.5cm 2.8cm 7.1cm},clip]{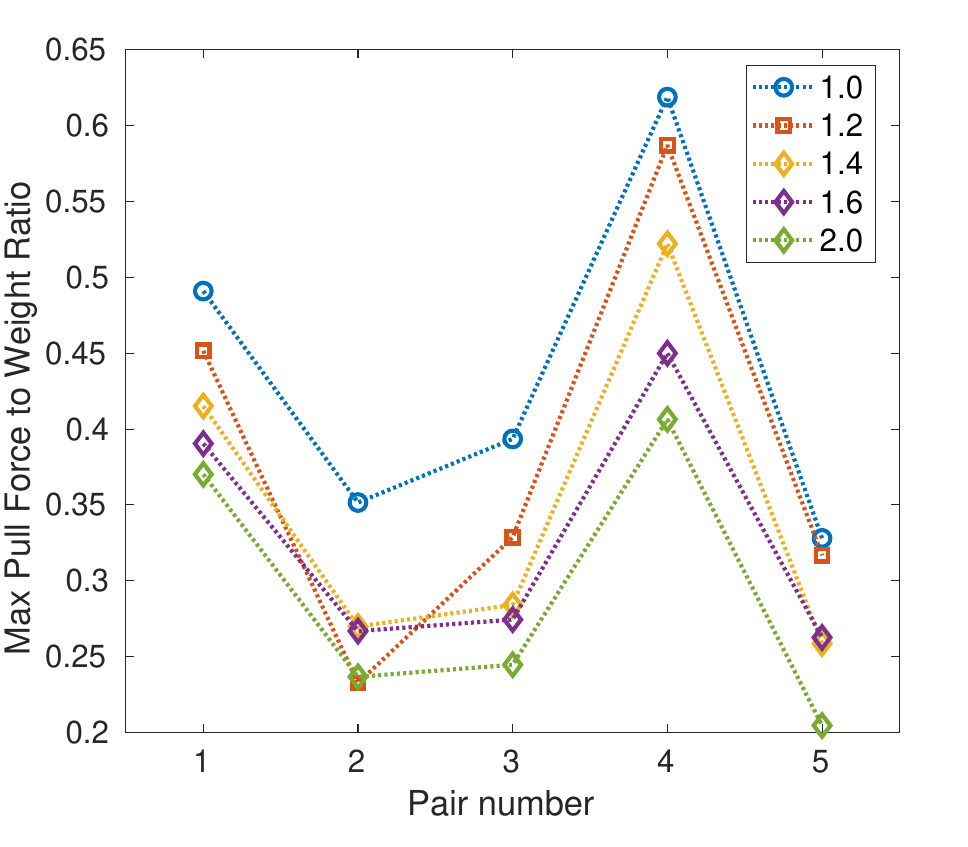}
          \vspace{1mm}\includegraphics[width=0.8\linewidth,trim={1.0cm 6.5cm 2cm 7.1cm},clip]{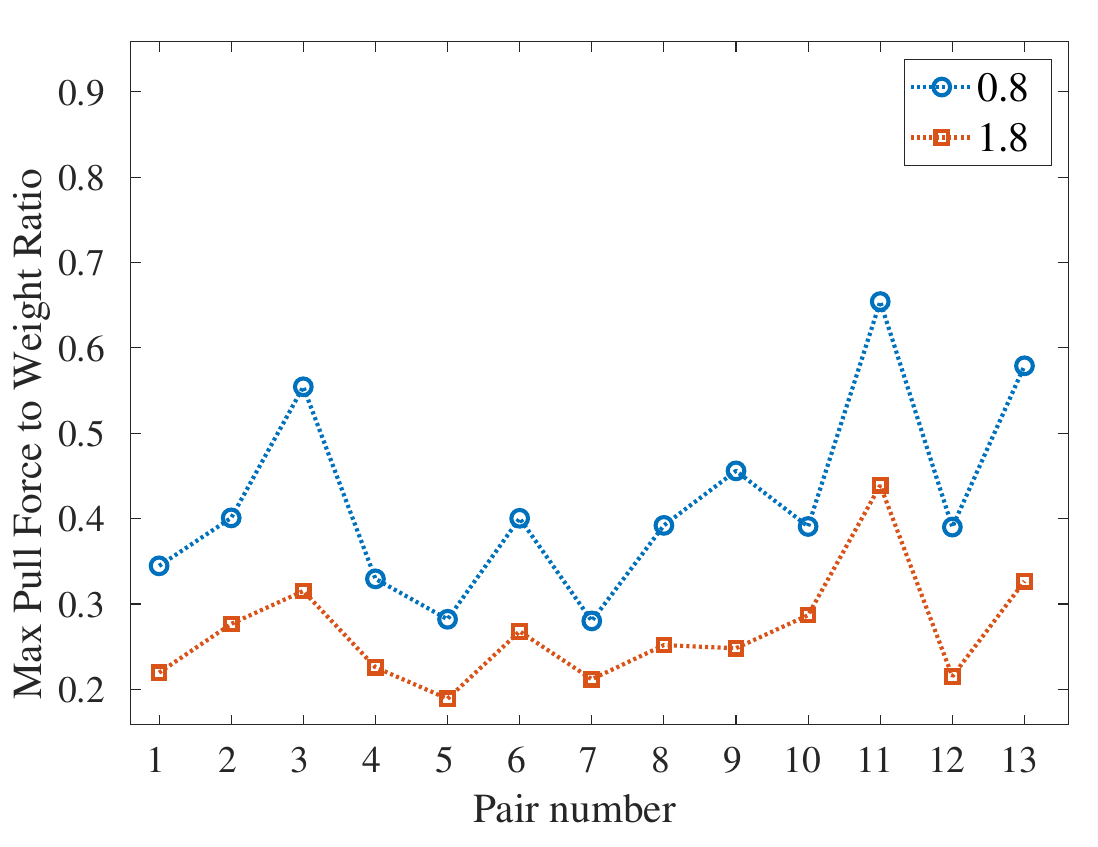}
  \caption{Mean of maximum pull force normalized to weight}
  \label{fig:max_pull_force_normalized_for_weight_new}
\end{figure}

In our experiment involving horizontal transfers, the pull force aligns with the direction of baton transfer. Thus, when the giver hands over the baton, the pull force is determined by the change in the $F_z$ component of forces measured by the interaction sensor (Fig. \ref{fig:intx_forces_handover}). We analyzed the maximum pull force during handovers by considering the greatest change in $F_z$ observed during the transfer time.

\begin{figure}[h!]
  \centering
      \setlength\abovecaptionskip{-0.3\baselineskip}      \includegraphics[width=0.8\linewidth,height=4.5cm,trim={1.0cm 6.5cm 2cm 6.5cm},clip]{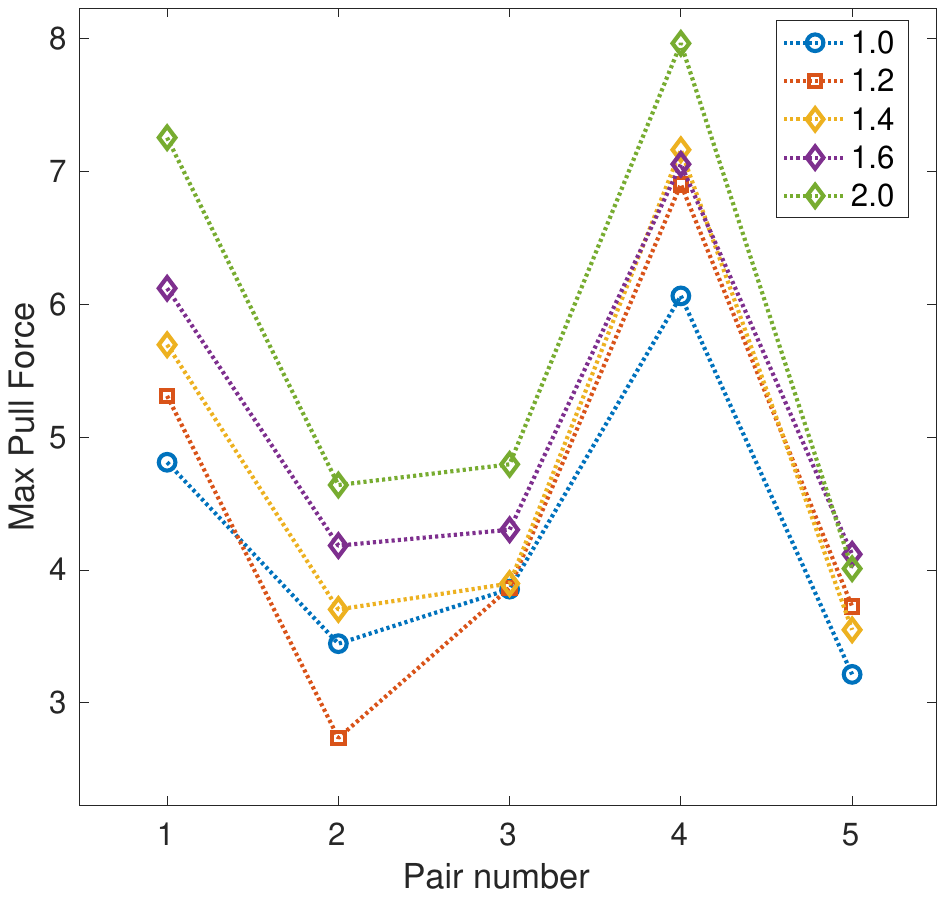}
      \includegraphics[width=0.8\linewidth,height=4.5cm,trim={1.0cm 6.5cm 2cm 7.1cm},clip]{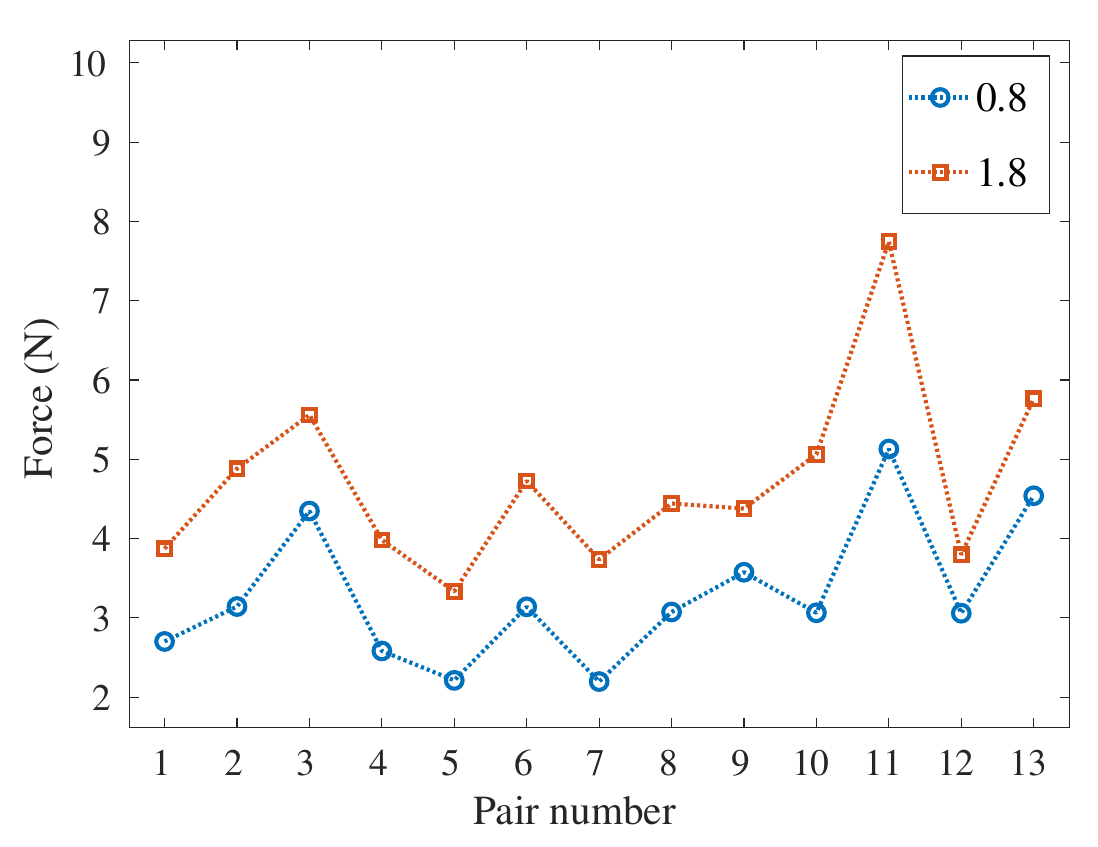}
  \caption{Mean of maximum pull force}
  \label{fig:max_pull_force}
\end{figure}

Considering the maximum pull force variation shown in Fig. \ref{fig:max_pull_force}, we observe that the average pull force is higher for heavier baton handovers than the lighter batons, with a significant increase for 1000g weight difference (p$<$0.05), inline with analysis the analysis of the previous dataset \cite{dataset-khanna}. Furthe, Fig. \ref{fig:max_pull_force_normalized_for_weight_new} shows the average maximum-pull force normalized to object weight across different settings. Similar to the analysis with the previous dataset in \cite{dataset-khanna}, we see that in general, this ratio decreases with an increase in the object weight, however not scaling linearlly with weight. 

\subsubsection{Interaction Forces - Load-sharing}
In our case of horizontal handovers, the felt load corresponds to the vertical component of interaction forces ($F_y$) in Fig. \ref{fig:intx_forces_handover}.
In our baton handovers, the load-share shift occurs when the measured $F_y$ changes its sign from negative to positive, at time: 
\begin{equation}
t_{ld\_shift} = min(t : F_y(t)>0 , t \in [t_{tak\_con},t_{giv\_rel}]). 
\nonumber
\end{equation}
\begin{figure}[h]
  \centering
  \includegraphics[width=0.8\linewidth,trim={0.8cm 5.2cm 2.3cm 5.9cm},clip]{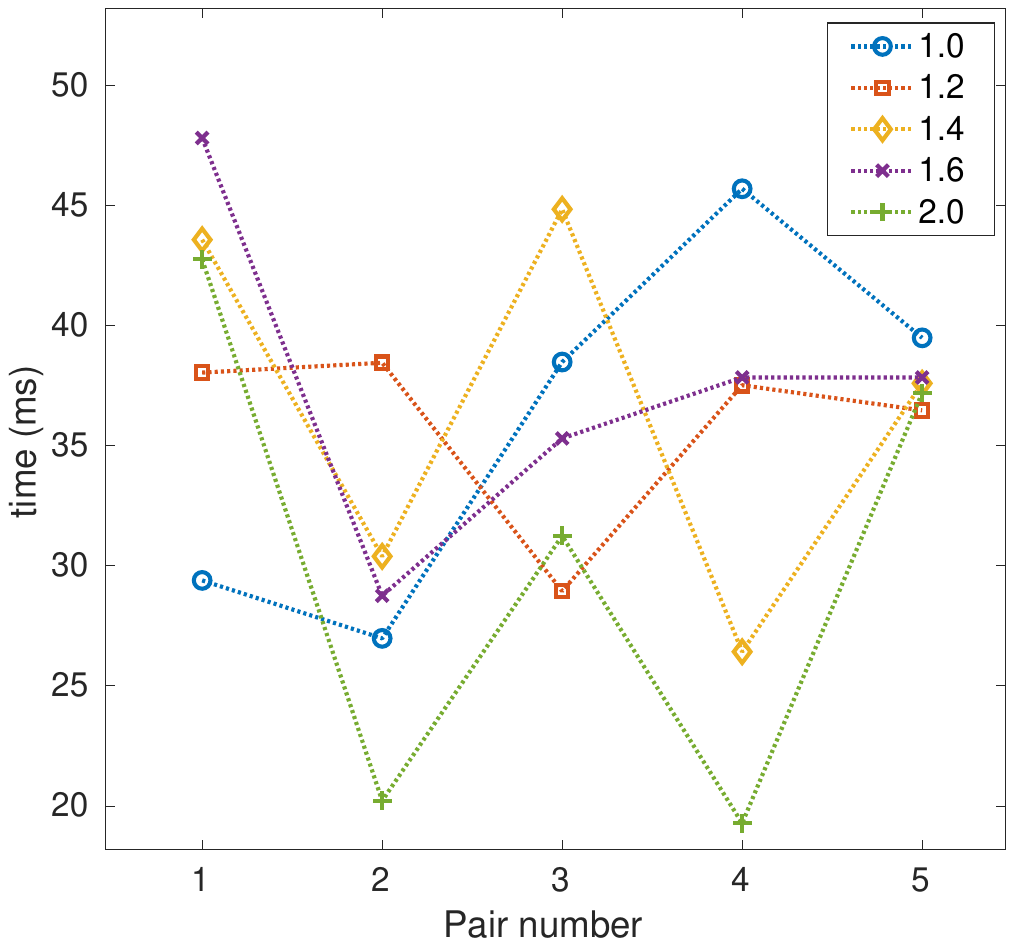}
    \includegraphics[width=0.8\linewidth,trim={0.8cm 6.8cm 2.3cm 5.9cm},clip]{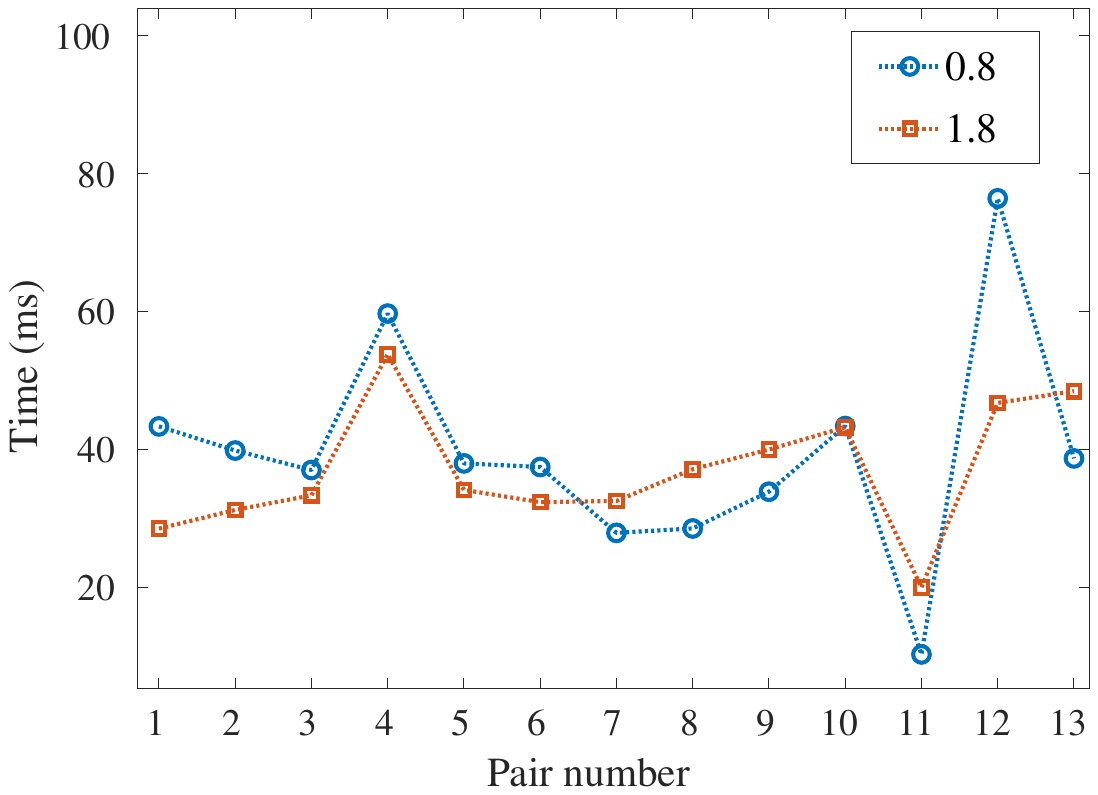}
  \caption{Mean time of load-share shift}
  \label{fig:tfull load from mid}
\end{figure}

We analysed if the load share shift occurs before or after the dominant grip force changes in the human handovers. 
In our datasets with the baton, the dominant grip force changes at the intersection point ($t=0$ ms) for each handover. 
Thus, we compare the load-share shift/transfer to the grip-force intersection, and the absolute $t_{ld\_shift}$ values
are sufficient for our analysis: a positive value means that the load-share shift occurs after the dominant grip force change at the intersection point. Figure \ref{fig:tfull load from mid} shows a plot of the mean $t_{ld\_shift}$ observed.

A positive mean is observed across the two weights, with values significantly above zero (p $<$ 0.05 in t-test analysis) for all pairs in both the datasets. This indicates that the load-share shift occurs after the intersection of grip forces for all object weights. In essence, this validates that the dominant grip force changes before the load-share shift is perceived by humans in handovers.

\subsection{Handover Motion Analysis}
Since motion capture (Mocap) data is prone to noise, we applied a Butterworth filter with a cutoff frequency of 5 Hz to smooth the collected motion data and reduce any high-frequency noise. This ensured cleaner and more reliable data for further analysis. It is important to note that the objects with added carefulness in handovers were excluded from the analysis presented in this work.
\subsubsection{\textbf{Height of Transfer}}
\begin{figure}[h]
  \centering
  \setlength\abovecaptionskip{-0.1\baselineskip}
  \includegraphics[width=0.8\linewidth,trim={0.2cm 5.7cm 2cm 6.6cm},clip]{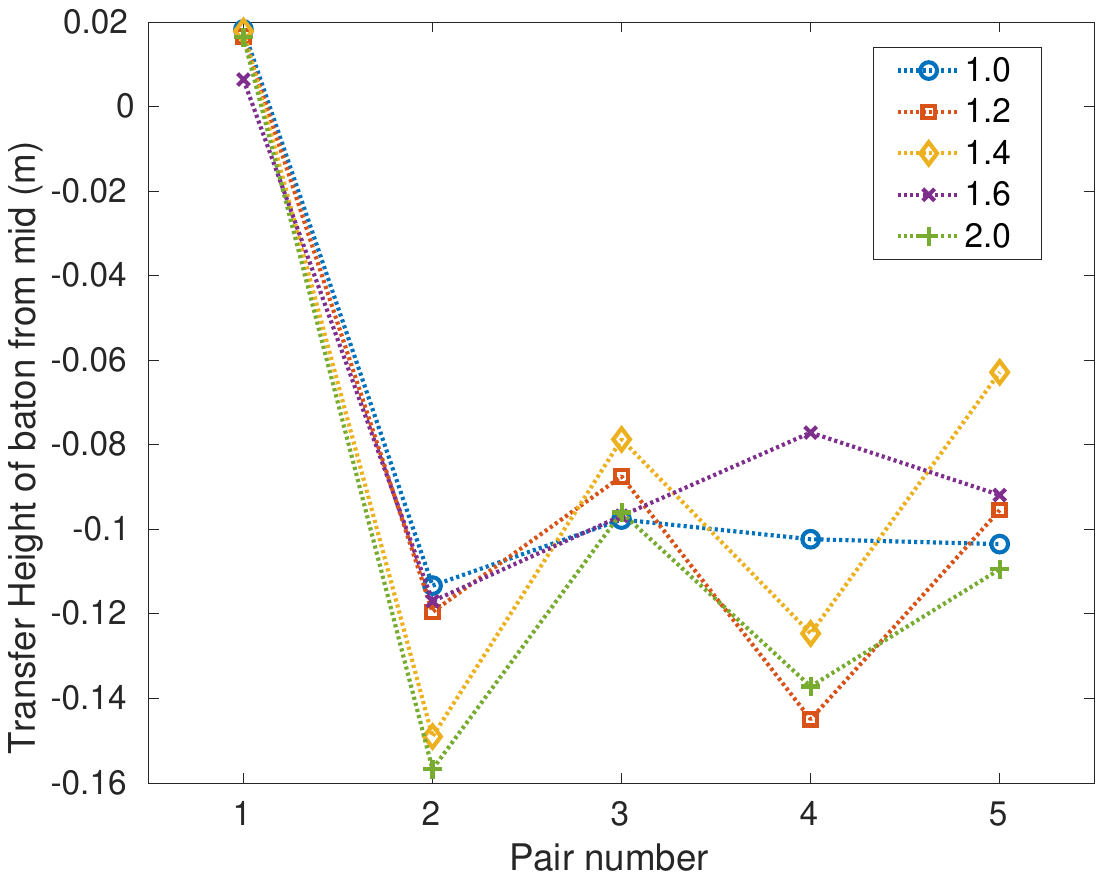}
  \includegraphics[width=0.8\linewidth,trim={0.2cm 5.7cm 2cm 6.6cm},clip]{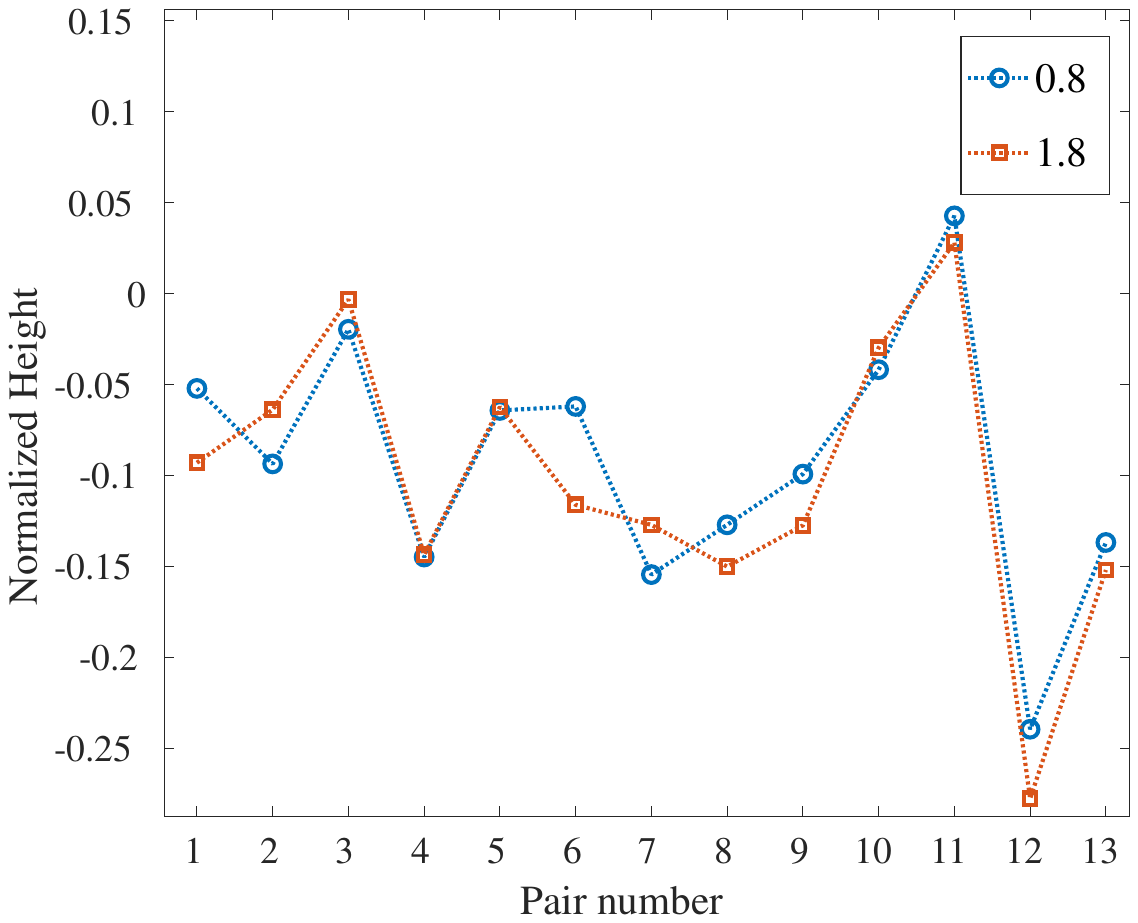}
  \caption{Mean normalized height of baton transfer}
  \label{fig:location_transfer_height}
\end{figure}
\begin{figure*}[t]
      \centering
        \includegraphics[width=\columnwidth,trim={0cm 0.0cm 0cm 0.3cm},clip]{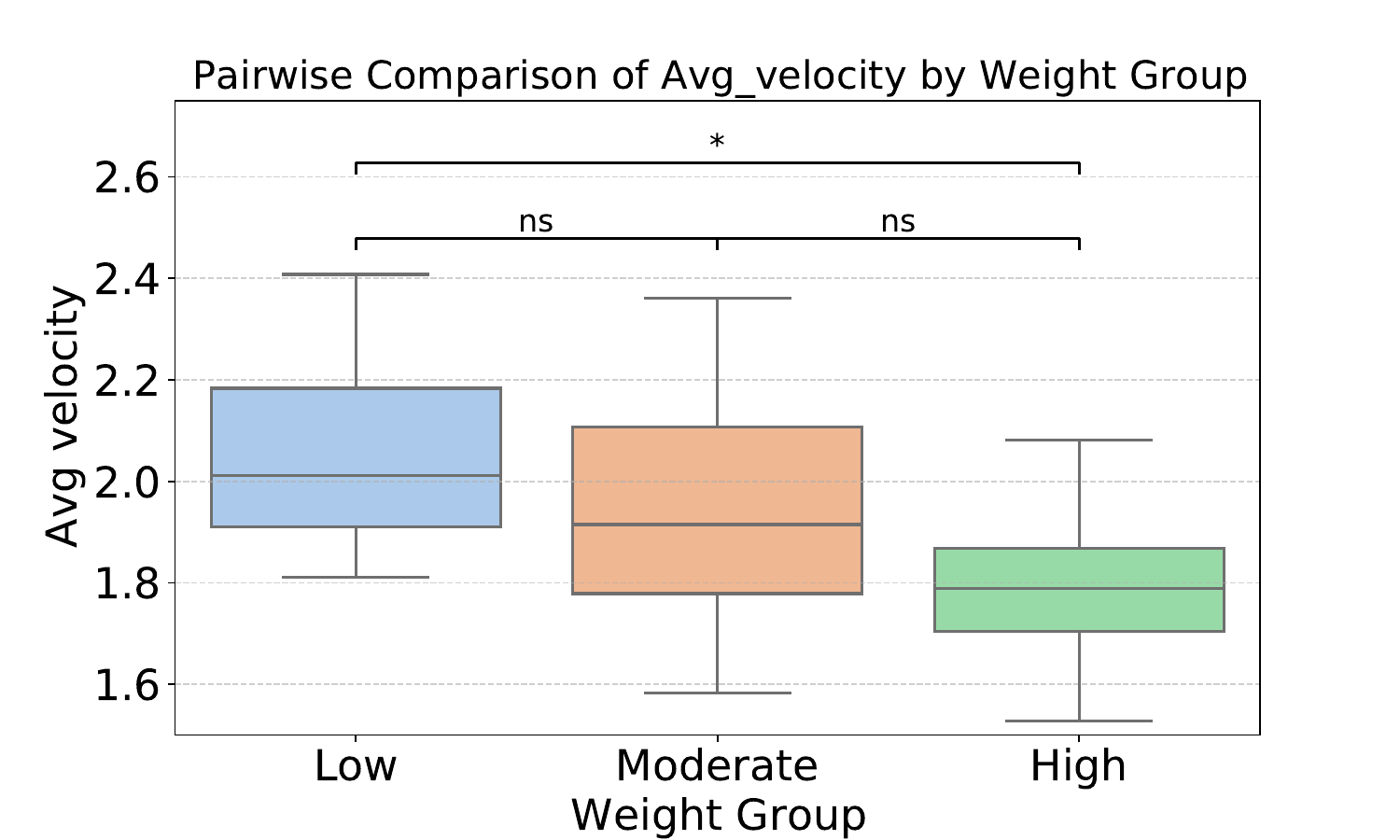}
        \includegraphics[width=\columnwidth,trim={0cm 0.0cm 0.3cm 0.3cm},clip]{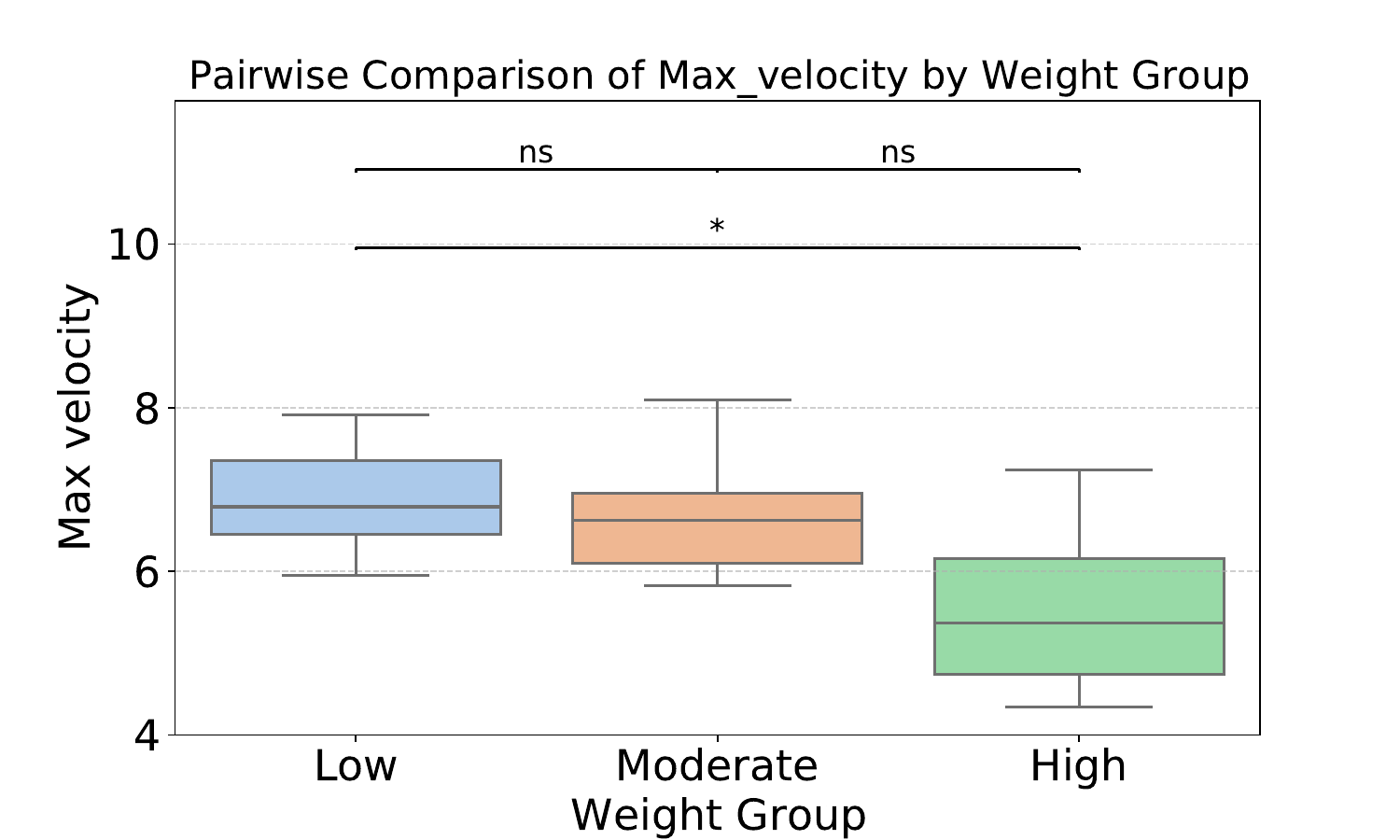}
        \vspace{1cm}
        \includegraphics[width=\columnwidth,trim={0cm 0.0cm 0.3cm 0.3cm},clip]{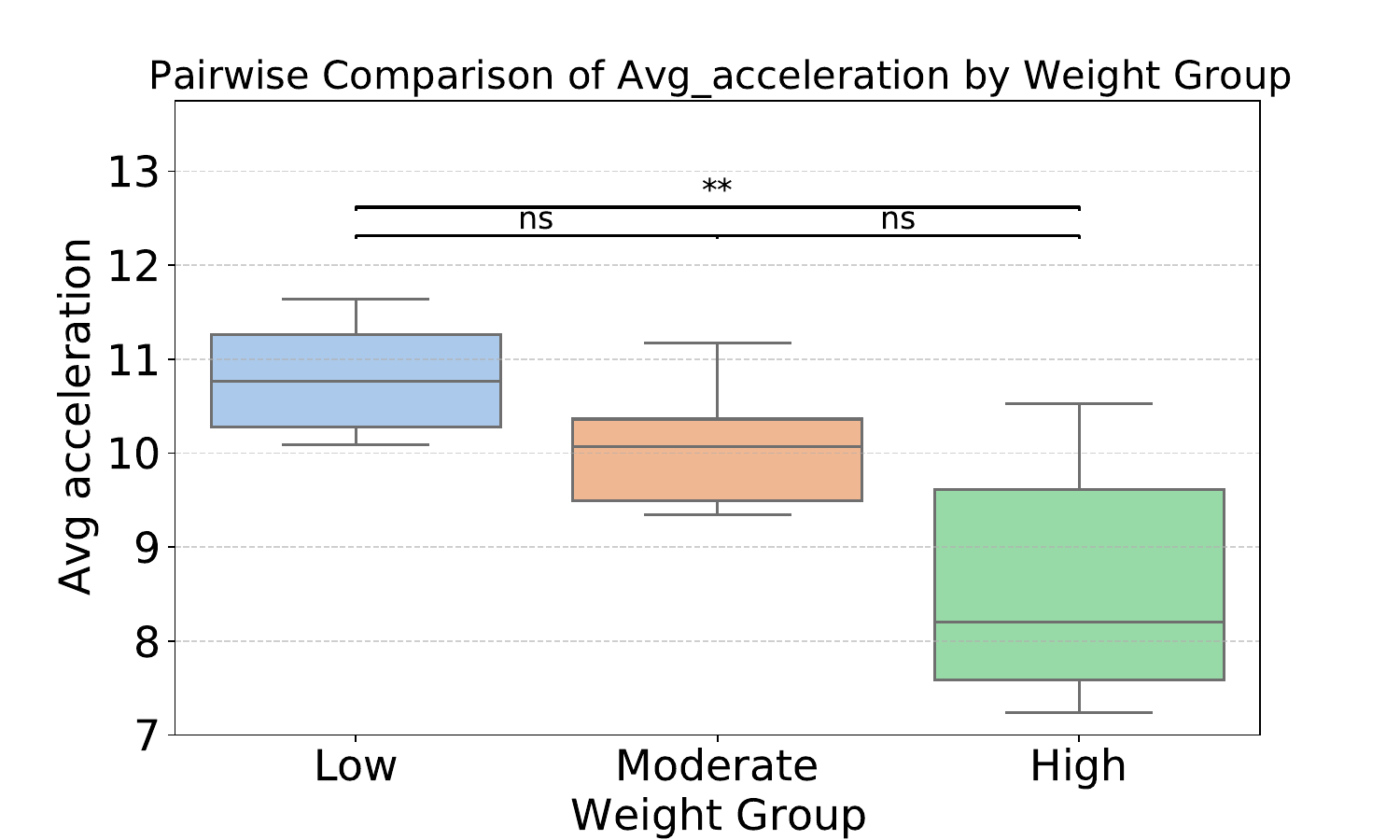}
        \includegraphics[width=\columnwidth,height=5.3cm,trim={0cm 0.0cm 0.1cm 0.3cm},clip]{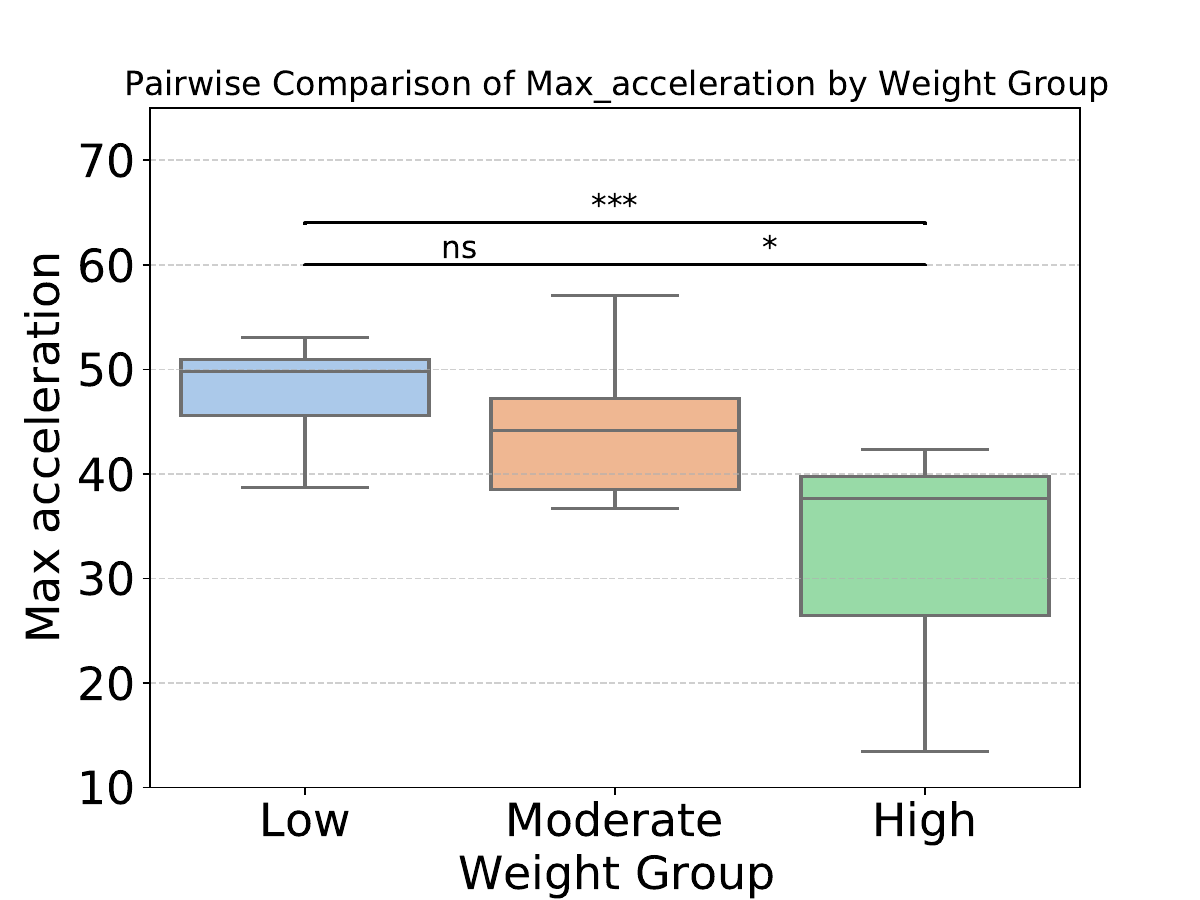}
    \setlength\abovecaptionskip{-2.4\baselineskip}
      \caption{Motion Characteristics across different Weight Categories}
      \label{fig:motion_box_plots_3categ}
   \end{figure*}
In \cite{h2H_handovers_how_dis_and_obj_mass_matter_Clint2017}, the analysis of a human-human study showed that transfer height does not depend on the object weight in handovers. This finding provides valuable insights into the natural dynamics of object transfer between individuals, suggesting a consistent spatial preference regardless of the item's weight.
Since objects shape and size could also influence the transfer height, we limit our analysis to the handover of objects with the same shape and size by using the New \textbf{Handovers@RPL-2.0} dataset with the sensor embedded baton handovers.
For all handovers, we consider the baton transfer height as the measured \textit{Z} coordinate of the baton at the intersection of grip forces, i.e., at $t=0$ ms. This precise moment represents the point at which both participants are equally engaged in supporting the object's weight. To ensure proper comparison across different participant pairs, this height was normalized to the average chest height of the giver and taker. This normalization accounts for variations in participant heights, allowing for a more standardized analysis.
The mean variation in the normalized height across different experimental settings is illustrated in Fig. \ref{fig:location_transfer_height}. 

Notably, our results indicate that there is no significant effect of object weight on transfer height, as also seen in our previous analysis \cite{dataset-khanna}, reinforcing the findings from previous studies. This suggests that the human behavior regarding transfer height during object handovers appears to remain consistent, regardless of the object's weight. 

\subsubsection{\textbf{Velocities and Acceleration}}
For this analysis, we consider the reaching phase motion of the human givers, i.e. the hand motion after picking the object and upto the object transfer in the handovers. We exclude the handovers with added

\begin{figure}[t]
      \centering
        \includegraphics[width=1.0\columnwidth,trim={0.5cm 0.5cm 0.5cm 0.3cm},clip]{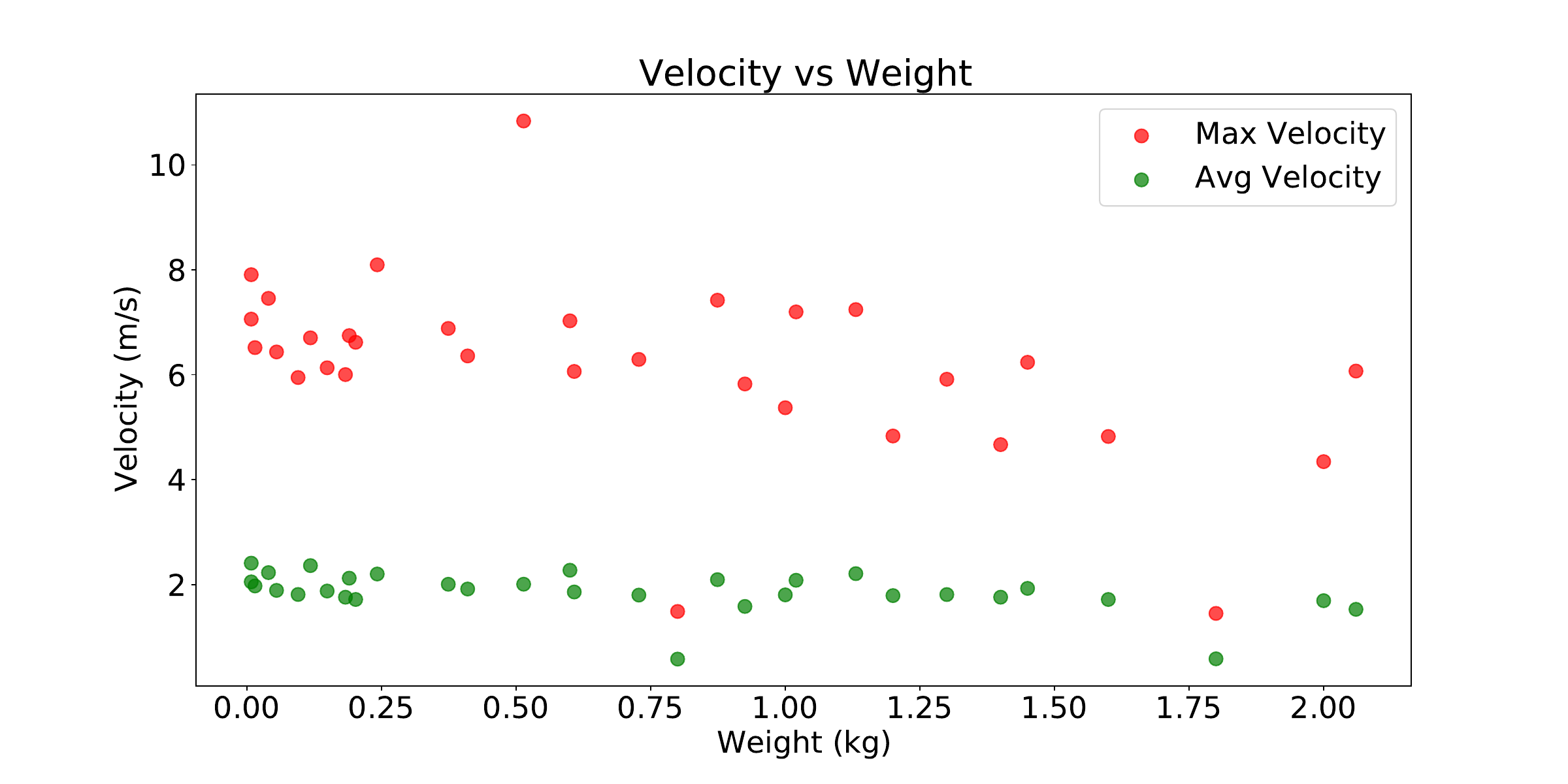}
        \vspace{1cm}
        \includegraphics[width=1.0\columnwidth,trim={0.5cm 0.5cm 0.5cm 0.3cm},clip]{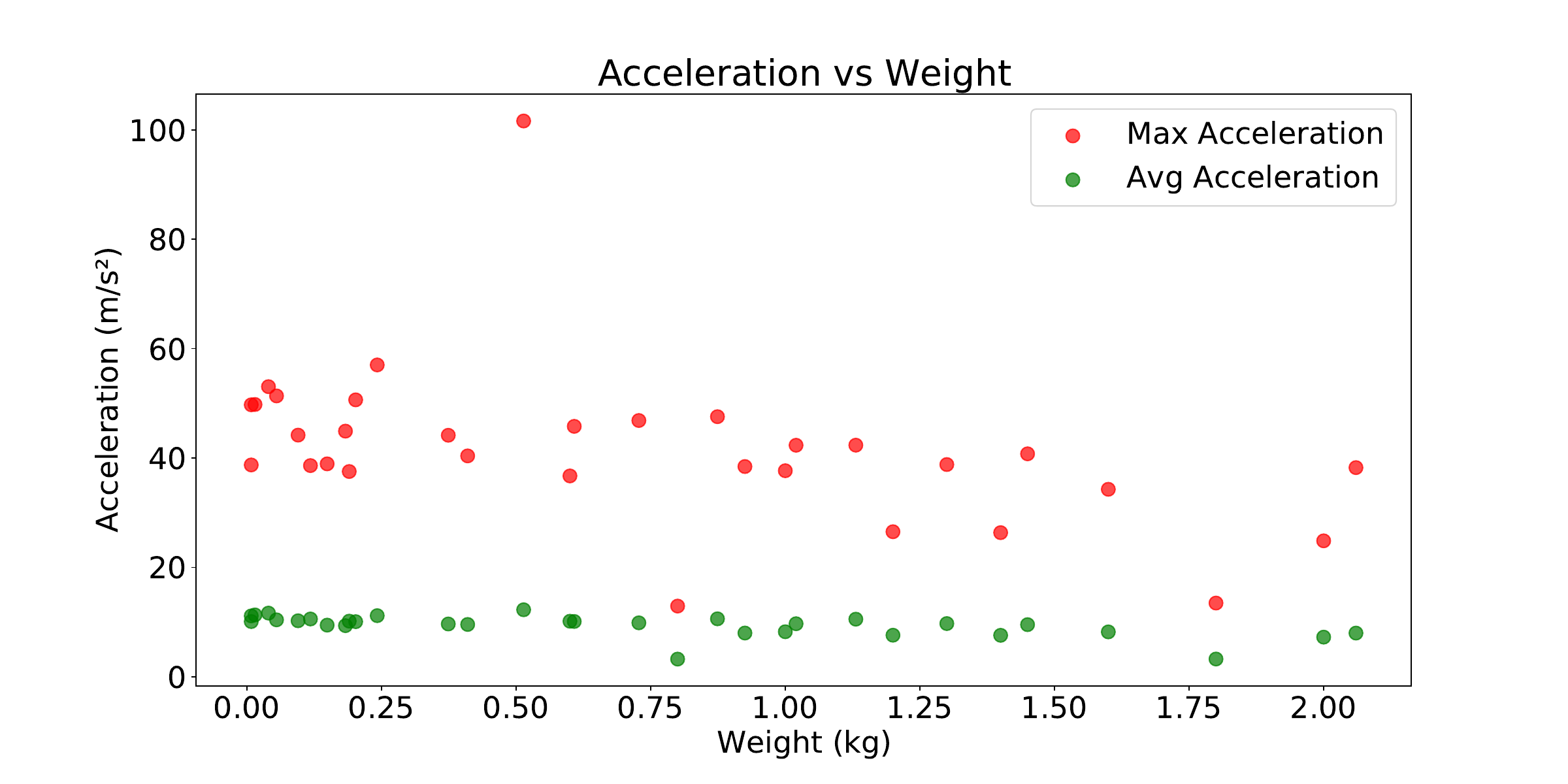}
    \setlength\abovecaptionskip{-2.4\baselineskip}
      \caption{Motion characteristics for human hand motion with different weighted objects in human-human handovers. Mean values observed for a particular object are plotted.}
      \label{figC:motion_avg_max_vel}
   \end{figure}

Fig. \ref{figC:motion_avg_max_vel} presents the mean values of average and maximum velocities, as well as average and maximum accelerations observed during handovers for each object across the three recorded datasets. This analysis focuses specifically on the movement of the human giver's hand carrying the object, as represented by 3D coordinates (position x, y, z). Objects requiring additional carefulness were excluded from this analysis.
A general trend emerges where these motion characteristics tend to decrease as object weight increases. However, several exceptions are noted, suggesting that factors beyond weight, such as object type, shape, and size, also play a role in influencing the velocity and acceleration of the human hand during handovers.

\textbf{Human hand motion Correlation with Object Weight}:

The relationship between object weight and motion characteristics (velocity and acceleration) was explored in the different recorded handover datsets, revealing different correlation patterns. 
Considering all recorded handovers across all object weights, our analysis reveals a negative correlation between motion characteristics and weight.

\begin{table}[h!] 
\centering
\caption{Correlation with Weight for Various Metrics}
\begin{tabular}{lcc} \hline & \textbf{weight} \\ \hline \textbf{avg\_velocity} & -0.473589 \\ \textbf{max\_velocity} & -0.488760 \\ \textbf{avg\_acceleration} & -0.603340 \\ \textbf{max\_acceleration} & -0.453666 \\ \hline 
\end{tabular} 
 \label{tab:correlation_with_weight} \end{table}
The correlations between weight and various metrics are summarized in Table~\ref{tab:correlation_with_weight}. A correlation coefficient (\(r\)) is classified as strong if \(|r| \geq 0.7\), moderate if \(0.5 \leq |r| < 0.7\), weak if \(0.3 \leq |r| < 0.5\), and very weak or negligible if \(|r| < 0.3\).
Based on this classification, the average acceleration (r = -0.603) shows a moderate negative correlation with weight, suggesting a moderate tendency for average acceleration to decrease as weight increases.
The correlations of weight with average velocity, maximum velocity, and maximum acceleration are all classified as weak negative correlations, though they are close to the moderate correlation threshold, indicating a weak tendency to decrease with increasing object weight.

Overall, while average acceleration demonstrates the strongest relationship with weight, these coefficients suggest that velocity and acceleration tend to decrease as object weight increases.
However, the absence of strong correlations indicates that the relationship is not robust or uniform across all participants.
This observation underscores the complexity of human handover dynamics and the potential influence of individual factors on motion characteristics during object handovers of varying weights.

For the previous dataset with sensor embedded object (\textit{Handovers@RPL}), i.e., the 0.8 kg and 1.8 kg baton handovers, we observed varied correlations between weight and motion across different pairs:

\begin{itemize}
\item \textbf{Moderate Positive Correlation:} Pairs 1, 5, 7, and 11 exhibited a moderate positive correlation between weight and motion. For these pairs, heavier weights were associated with higher velocities (both average and maximum).

\item \textbf{Weak Positive or Neutral Correlation:} Pairs 2 and 4 showed weak or inconsistent correlations between weight and motion, with little to no effect on velocity and acceleration.

\item \textbf{Moderate to Strong Negative Correlation:} Pairs 3, 6, 8, 9, 10, 12, 13 demonstrated a moderate or strong negative correlation where heavier weights corresponded to lower velocities.
\end{itemize}
These diverse correlations suggest that the relationship between object weight and handover motion varies significantly among different pairs, indicating potential influences of individual strategies or physical characteristics on handover dynamics.

Finally, for the new data of handovers with sensor-embedded baton (\textit{Handovers@RPL-2.0}), i.e. object weighing 1000g, 1200g, 1400g, 1600g, and 2000g, we observed varied correlations between weight and motion characteristics (velocity and acceleration), categorized as follows:

\begin{itemize}
\item \textbf{Weak Negative Weight Correlation:} Only Pair 1 exhibited a weak positive correlation across all motion metrics, with average velocity showing a correlation of \(-0.532\), maximum velocity at \(-0.682\), average acceleration at \(-0.387\), and maximum acceleration at \(-0.593\).
This suggests that for this pair, increasing weight slightly decreases both velocities and accelerations. However, the observed negative correlations are not as pronounced compared to other pairs.

\item \textbf{Strong Negative Weight Correlation:} All other pairs exhibited strong negative correlations between weight and motion metrics. For example, Pair 2 showed a significant negative correlation with average and maximum velocities at \(-0.645\) and \(-0.642\), respectively, and average and maximum accelerations at \(-0.681\) and \(-0.642\).  
On average, these pairs demonstrated correlations of \(-0.650\) for average velocity, \(-0.309\) for maximum velocity, \(-0.570\) for average acceleration, and \(-0.286\) for maximum acceleration, confirming that as weight increases, both velocities and accelerations tend to decrease across the pairs.

\end{itemize}

Thus, weight generally tends to have a strong negative correlation with most motion metrics, indicating that heavier weights typically result in decreased velocities and accelerations.  
However, this relationship can vary across individuals. In certain instances, an increase in weight may lead to more dynamic motion, resulting in a positive correlation with motion metrics. Such cases suggest that the effect of weight on motion may not always be linear. Some individuals may exhibit increased velocity and acceleration with heavier objects due to their unique handling strategies or compensatory mechanisms. These findings highlight the significant role of individual differences in handover dynamics.

\textbf{Weight Categories}:
For further analysis, we categorized all objects in Table \ref{tab:handover_summary} into three weight categories: Low, Moderate, and High. 

We explored different weight limits for the three categories to find clear and significant differences in motion characteristics, particularly in velocities and accelerations. This approach helped us identify meaningful distinctions between the three weight categories, providing a solid foundation for our further analysis.
The box-plots in Fig. \ref{fig:motion_box_plots_3categ} shows the comparison of different motion metrics for reaching motion in human handovers the three categories of objects, with obsereved significant differences in some of these metrics.
For Low vs High weight categories, we found significant differences in the maximum velocity (\(p = 0.027\)), average acceleration (\(p = 0.002\)), and maximum acceleration (\(p = 0.004\)). Additionally, for Moderate vs High categories, we found significant differences average acceleration (\(p = 0.033\)) and maximum acceleration (\(p = 0.014\)). These results show clear and consistent differences in velocity and acceleration for the human handover motion for different categories of objects with increasing weights.

\begin{table}[h!]
\centering
\caption{Average and Maximum Velocities and Accelerations by Weight Group}
\resizebox{0.995\columnwidth}{!}{
\begin{tabular}{lcccc}
\hline
\textbf{weight\_group} & \textbf{avg\_velocity} & \textbf{max\_velocity} & \textbf{avg\_acceleration} & \textbf{max\_acceleration} \\
\hline
Low & 2.023 & 6.816 & 10.592 & 47.445 \\
Moderate & 1.840 & 6.506 & 9.404 & 45.154 \\
High & 1.680 & 5.221 & 7.928 & 32.951 \\
\hline
\textbf{Avg weight} & 0.0384 & 0.428 & 1.457 \\
\hline
\end{tabular}
}
\label{tabT:vel_acc_weights}
\end{table}

The following summarizes the key findings for different weight categories:
\begin{itemize}
    \item \textbf{Velocity Trends:}
    \begin{itemize}
        \item Low-weight objects (8--100g) achieve the highest average velocity (2.02 m/s) and maximum velocity (6.81 m/s).
        \item High-weight objects (950--2060g) show the lowest values for both average velocity (1.68 m/s) and maximum velocity (5.22 m/s).
        \item Moderate-weight objects (100--950g) exhibit intermediate velocities, slightly higher than the High group but significantly lower than the Low-weight group, with an average velocity of 1.84 m/s and a maximum velocity of 6.51 m/s.
    \end{itemize}
    
    \item \textbf{Acceleration Trends:}
    \begin{itemize}
        \item Low-weight objects exhibit the highest average acceleration (10.59 m/s\(^2\)) and maximum acceleration (47.44 m/s\(^2\)).
        \item High-weight objects show the lowest values for both average acceleration (7.93 m/s\(^2\)) and maximum acceleration (32.95 m/s\(^2\)).
        \item Moderate-weight objects display intermediate acceleration values, higher than the High group but lower than Low-weight objects, with an average acceleration of 9.40 m/s\(^2\) and maximum acceleration of 45.15 m/s\(^2\).
    \end{itemize}
    
    \item \textbf{Overall Patterns:}
    \begin{itemize}
        \item Both velocity and acceleration tend to decrease as weight increases.
        \item The High-weight group shows the most pronounced impact of weight, causing the lowest values for both velocity and acceleration.
    \end{itemize}
\end{itemize}

\section{Adapting Robotic Motion to Object Weight}
\label{section:adaptive_motion}
We propose using the insights from our analysis to command changes in robotic motion based on object weight and test this in robot-to-human handover experiments.
In these experiments, we aimed to evaluate how a robot giver can convey an object's weight through its handover motion to a human taker. The objects used for these experiments were drawn from the three weight categories defined in our analysis: Low (0.008–0.1 kg), Moderate (0.1–0.95 kg), and High (0.95–2.06 kg).
To ensure that participants could not visually identify weight differences, we used identical-looking boxes for all three weight categories, adjusting their weights to represent each group. The Low weight category was represented by the box at its base weight of 0.09kg (90g). For the Moderate and High categories, additional weights were added to the boxes, bringing them to 0.9 kg and 1.180 kg, respectively. The upper weight limit was constrained by the payload capacity of the Baxter robot, which was used for the handover tasks.
Thus, the robot performed the handovers using these identical-looking boxes while adapting its motion to the object weight, drawing inspiration from the dynamics of human-to-human handovers.
\subsection{Proposed Adaptive Motion}
We started with a fixed trajectory of waypoints for the handover motion of a robot giver, the Baxter robot, to hand over the object to a human taker.
This trajectory and the experimental setup was similar to that obsereved in our human-human handovers: a human giver taking the object from the corner of a table and handing the object mid-air over the center of the table to a human taker. The robot's right arm was used for this experimentation and this trajectory was kept the same for both left and right handed participants.
We aimed for a minimum-jerk trajectory for the robotic arm. Our analysis revealed that the maximum acceleration varied the most among the three weight categories in human handover motion. Thus, we adjusted the maximum acceleration for the robot trajectory for the three weight categories, using the aforementioned identical-looking boxes. However, the robot motion is limited by its own safety constraints. To account for these variations, we set the robot's maximum acceleration to 20.575 m/s\(^2\) for High  weight, 27.575 m/s\(^2\) for Moderate weight, and 38.676 m/s\(^2\) for Low weight objects, ensuring that the robot's motion remained within safe limits while accommodating the different weight categories. The corresponding average accelerations for the robot handover trajectory (to give an object) were found to be 4.580 m/s\(^2\) for High weight, 4.920 m/s\(^2\) for Moderate weight, and 5.270 m/s\(^2\) for Low weight.

\subsection{Baseline}
We will evaluate this adaptive motion in human-robot handover experiments against a baseline motion that follows a the same motion trajectory with acceleration values for the Moderate weight category. This comparison will help us assess the effectiveness of our adaptive strategy in enhancing the naturalness and efficiency of robot-human handovers across different object weights.

\subsection{Experimentation}
\begin{figure}[h!]
      \centering
     \includegraphics[width=0.9\columnwidth,trim={0.5cm 0.6cm 0.0cm 2.880cm},clip]{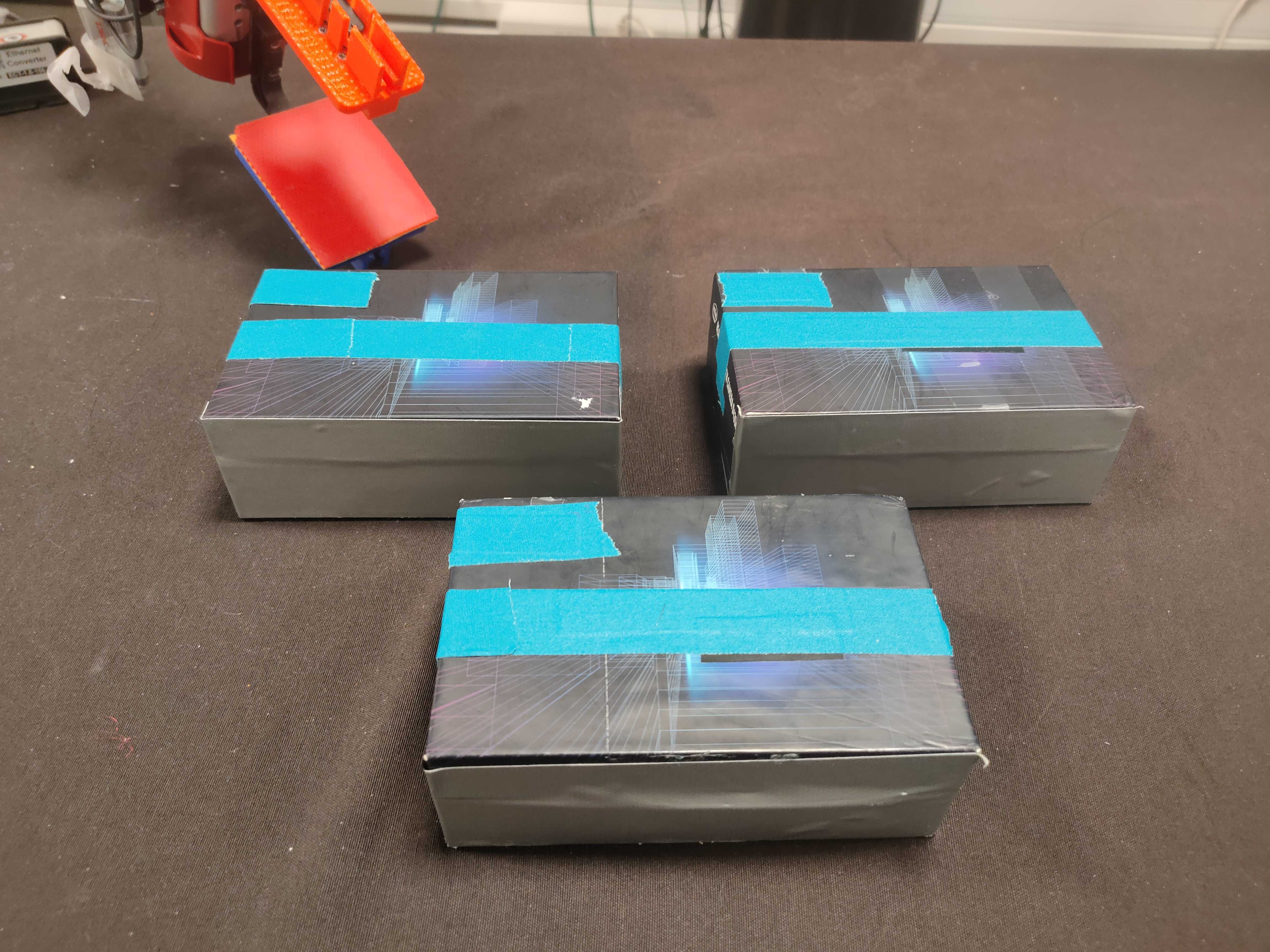}  
      \caption{Identical-looking objects of different weights used for evaluation in front of the robot gripper.
      The center object weighs 1.18 kg, the left object weighs 0.09 kg, and the right object weighs 0.9 kg.}
      \label{fig:Objects_evaluation_new_handovers}
   \end{figure}

We performed robot-to-human handovers with the Baxter robot handing over objects to participants. We evaluated the adaptive motion against the baseline motion by having the Baxter robot perform two handovers for each of the adaptive and baseline motions, labeled as strategies X and  Y, for each of the three objects. 
The participants had no prior knowledge of the object weights. To prevent them from guessing the object weights, the participants did not see the robot picking up the objects (hidden from their field of view). 
The participants were instructed ot focus on the handover motion of the robot as it gives them the object. 
Once the robot had the object in its gripper, the participants see the robot at the same starting pose for each object. Then, the  robot would execute the desired handover motion to reach the handover pose. The grip release was commanded to be a 
automatic timed grip release, i.e. the robot would open its gripper once it reached the handover pose. Owing to a big gripper base, the object doent fall from the gripper and the participants were instructed to take the object only after the robot has opened its gripper.

For evaluation, we used subjective measures in the form of questionnaires. After encountrering a particular  where baxter handed them the 3 objects two times each in a random order, the participants were asked to rate their experience on a 1-7 scale for various questions about the robot’s motion. The questions were as follows:

\begin{enumerate}
    \item How natural did the handover feel? (1 Not natural at all - 7 Very Natural)
    \item How well could you perceive differences in the robot's motion for different objects? (1 Not at all - 7 Very well)
    \item How well could you anticipate the object's weight before grasping it? (1 Not at all - 7 Very well)
    \item How helpful was the robot's motion in anticipating the object's weight? (1 Not helpful at all - 7 Very helpful)
    \item How well could you prepare for receiving objects of different weights? (1 Not at all - 7 Very well)
    \item How safe did the handover feel in this strategy? (1 Not safe at all - 7 Very safe)
    \item How confident did you feel receiving the objects in this strategy? (1 Not at all - 7 Very well)
\end{enumerate}

After the participants had encountered both strategies, they answered:
    \begin{enumerate}
    \item Which strategy did you prefer overall for receiving objects from the robot? (X, Y or other answer)
    \item Which strategy made it easier to estimate the object's weight? (X, Y or other answer)
    \item In which strategy did you feel more confident receiving the objects? (X, Y or other answer)
    \item Which was the most natural out of the two? (X, Y or other answer)
    \item Can you guess the object weights? (for the 3 objects)
\end{enumerate}

\textbf{Hypotheses:} We hypothesize that adaptive motion in robot-to-human handovers is perceived as better than non-adaptive motion for objects of different weights on the following aspects:  
\begin{itemize}
    \item \textbf{HM1:} Naturalness of the handover.
    \item \textbf{HM2:} Perceived differences in the robot's motion.
    \item \textbf{HM3:} Ability to anticipate object weight.
    \item \textbf{HM4:} Helpfulness of the robot's motion in weight anticipation.
    \item \textbf{HM5:} Preparedness for receiving objects of different weights.
    \item \textbf{HM6:} Perceived safety of the handover.
    \item \textbf{HM7:} Confidence in receiving the object.
\end{itemize}

\subsection{Results}
A total of 20 participants took part in this robot-to-human handovers' experimentation.
Based on subjective responses, we found that most participants preferred the adaptive motion strategy over the baseline for receiving objects, describing it as more natural. The majority also reported that it was easier to estimate the object's weight with the adaptive strategy, and they felt more confident with this strategy compared to the baseline.
\begin{figure}[h!]
      \centering
       \includegraphics[width=0.9\linewidth,trim={1cm 8.5cm 1.8cm 8.3cm},clip]{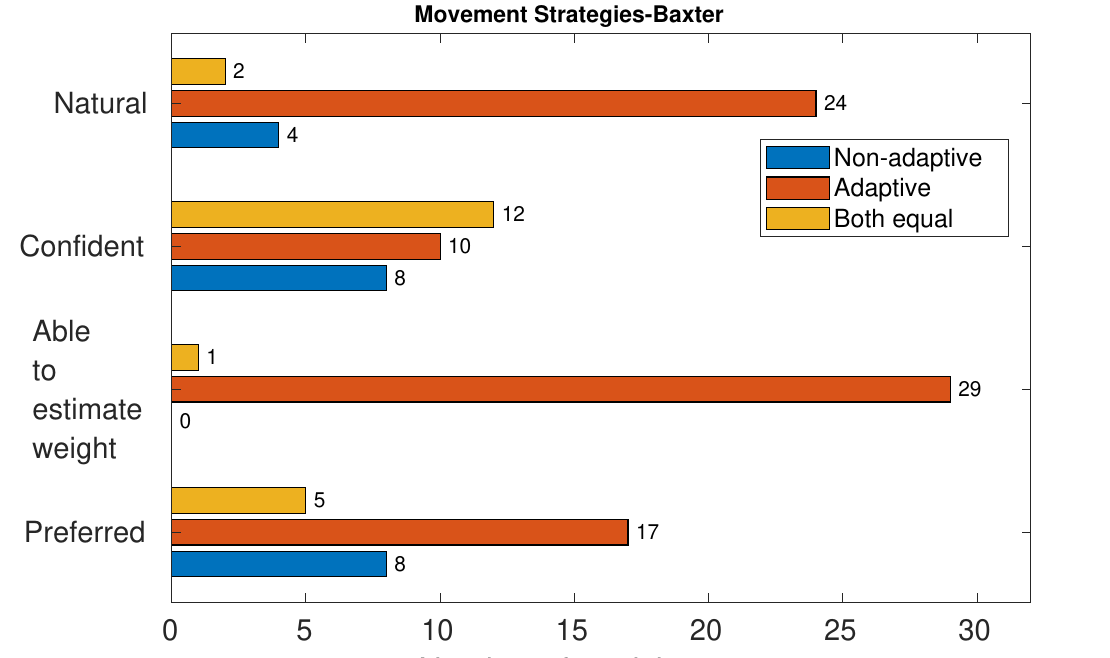}
    \caption{Responses comparing different movement strategies}
      \label{figT:Answers_move_Baxter}
   \end{figure}

\begin{figure}[t]
      \centering
       \includegraphics[width=0.9\linewidth,trim={0.2cm 3.5cm 0.2cm 3.3cm},clip]{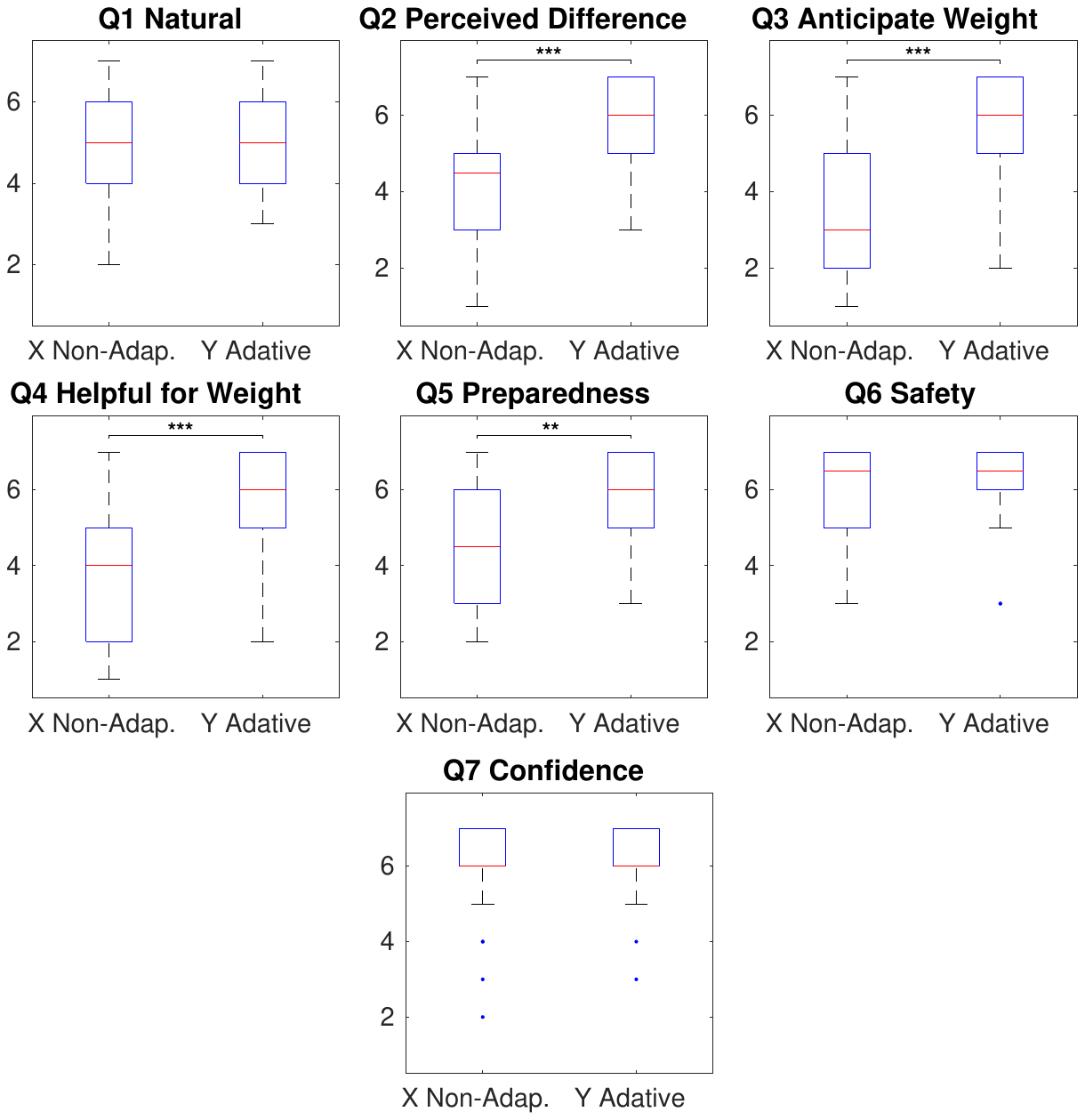}
    \caption{Participants' Subjective Responses comparing movement strategies}
      \label{fig:Answers_movemnt_strategies}
   \end{figure}

This suggests that human receivers were better able to estimate the object weight from the adaptive motion, which was inspired by human-human handovers, even though it was not an exact replica of the changes in velocity and acceleration observed in human motion.

Furthermore, all participants were able to identify at least two weight categories—a lightweight object and a heavy object—indicating that the difference in motion was most noticeable between these two weight extremes. This highlights the effectiveness of the adaptive strategy in conveying weight information to a human taker via motion of the robot giver in robot-to-human handovers.
Considering questions 1-7, the results from the pairwise t-tests indicate that the adaptive motion (Strategy-Y) outperforms the non-adaptive motion (Strategy-X) in most areas, with significant improvements in the following categories:

\begin{itemize}
    \item \textbf{Perception of robot's motion for different objects:} Strategy-Y shows a significant improvement over Strategy-X ($t = -4.705, p < 0.001$). \textbf{HM2} is found valid.
    \item \textbf{Anticipation of object weight before grasping:} Strategy-Y significantly outperforms Strategy-X ($t = -4.721, p < 0.000$).  \textbf{HM3} is found valid.
    \item \textbf{Helpfulness of robot's motion in anticipating object weight:} Again, Strategy-Y shows a significant improvement ($t = -4.527, p < 0.001$).  \textbf{HM4} is found valid.
    \item \textbf{Preparation for receiving objects of different weights:} Strategy-Y leads with a significant difference ($t = -3.231, p = 0.002$).  \textbf{HM5} is found valid.
\end{itemize}

However, there is no significant difference between the two motion strategies in the following categories:

\begin{itemize}
    \item \textbf{How natural the handover felt:} No significant difference ($t = 0.095, p = 0.925$).  Thus, even though adaptive motion was scored better, \textbf{HM1} is not found valid. 
    \item \textbf{How safe the handover felt:} No significant difference ($t = -0.717, p = 0.476$). \textbf{HM6} is found not valid.
    \item \textbf{How confident the user felt receiving the objects:} No significant difference ($t = -0.791, p = 0.433$). \textbf{HM7} is found not valid.
\end{itemize}

In conclusion, the adaptive motion (strategy Y) significantly improves the user experience in perceiving motion differences, anticipating object weight, and preparing for weight differences, but it does not show significant differences in the perceived naturalness, safety, or confidence during the handover process.

Overall, our findings highlight the effectiveness of the adaptive strategy in conveying weight information to a human receiver via the robot's motion in robot-to-human handovers. However, further research and experimentation are needed to make the weight-adaptive robot motion more natural in robot-to-human handovers.

\section{Learning Grip Release : Strategy Adjusted to Weight of object}

We propose a human-inspired data-driven strategy that adjusts to object weight for commanding robotic grip release in robot-to-human handovers. This approach leverages the diverse weight classes in our dataset with the handover forces to develop a more adaptive and robust grip-release strategy. To evaluate the effectiveness of this strategy and perform a quantitative assessment of the dataset, we implement and test the proposed strategy across different object weights in robot-to-human handover scenarios.

\subsection{Data-driven strategy}
\subsubsection{Time series Formulation}
For each recorded handover, the interaction wrench $W_{int}$ comprises of a time series of 6 F/T components that vary significantly in the handover, as seen for forces in Fig. \ref{fig:intx_forces_handover}. As seen in Fig. \ref{fig:forces_handover}, the giver's grip forces start to decline before the intersection point ($t=0$ ms). The learning task is to predict the grip release start from the observed $W_{int}$ before $t=0$ ms. Thereby, we sampled a time series of loadshare force ($F_y$ in Fig. \ref{fig:intx_forces_handover}), pull force ($F_z$) from $W_{int}$ and the weight of the object ($w$), ending at a time step $t_e$ from each saved handover:
 \begin{equation}
 X(t_e)= \{F_{y}^{t}, F_{z}^{t}, w : t \in [t_e\text{-} 100,t_e]\}.
 \label{eqn:time_series_Int_wrench}
 \end{equation} 
This 100-step time series corresponds to 833 ms at 120 Hz for our recorded data, with $t_e$ varying between 0 to -215 time steps (-1803 ms).

We use $t_{giv\_rel}$ to assign labels to the time series $X(t_e)$:

 \begin{equation}
    y(X(t_e))=\begin{cases}
    1,& \text{if } t_e>=t_{rel\_start}\\
    0,& \text{if } t_e<t_{rel\_start} 
    \end{cases}
    \label{eqn:Label creation te}
 \end{equation}
An LSTM classifier assigns binary labels (1 or 0) to indicate whether the input $X(t_e)$ corresponds to the initiation of grip release in human handovers. A robotic giver can leverage this approach by measuring interaction forces during a handover and utilizing a similar classifier, trained on human data, to initiate a human-inspired grip release. This method enables the robot to emulate human-like behavior in determining the appropriate moment to release its grip during object handovers.

The labeled samples were then used to train a long short-term memory (LSTM) based classifier. The trained LSTM is then used for robot to human handovers where $W_{int}$ is measured.
Once the end pose for handover is reached, the $W_{int}$ time series is input to the classifier for commanding the robotic grip-release.
The LSTM classifier assigns the labels 1 and 0 implying whether the input $W_{int}$ corresponded to a grip release start in human handovers.
\subsubsection{LSTM Architecture} 
In our previous work formulating a data-driven grip release strategy \cite{parag-humanoids}, we employed a classic LSTM architecture. However, for the current study, we have opted for a VAE-LSTM architecture to enhance the strategy's adaptability to varying object weights. This advanced model combines the strengths of Variational Autoencoders (VAE) and Long Short-Term Memory (LSTM) networks, allowing for more robust and flexible predictions across a range of object masses. The VAE component enables the model to learn a latent representation of the input data, potentially capturing underlying patterns related to weight variations, while the LSTM component maintains the ability to process and predict based on temporal sequences of interaction forces. This change aims to improve how well our grip release strategy works across various weights in real-world handover situations.


\begin{figure}[h]
  \begin{tikzpicture}[
    > = Stealth,
    process/.style = {rectangle, draw, text width=3cm, align=center, minimum height=1cm},
    lstm/.style = {rectangle, draw, text width=3cm, align=center, minimum height=1cm, fill=blue!20},
    connection/.style = {->, thick},
    latent/.style = {ellipse, draw, text width=2.5cm, align=center, minimum height=1cm, fill=green!20},
    output/.style = {rectangle, draw, text width=3cm, align=center, minimum height=1cm, fill=orange!20}
]

\node[process] (input) {Input Sequence ($X(t_e)$)};
\node[lstm, below=of input] (encoder) {Encoder LSTM};
\node[latent, below=of encoder] (latent) {Latent Space};
\node[latent, below right=0.05cm and 2cm of latent] (sampling) {Reparameterize (Sampling)};
\node[lstm, below=of latent] (decoder) {Decoder LSTM};
\node[output, below=of decoder] (sigmoid) {Sigmoid Activation};
\node[output, below=of decoder, xshift=3.5cm, text width=2.5cm, align=center, fill=orange!20] (output) {Binary Output ($p$)};

\draw[connection] (input) -- (encoder);
\draw[connection] (encoder) -- (latent);
\draw[connection] (latent) -- (sampling);
\draw[connection] (sampling) -- (decoder);
\draw[connection] (decoder) -- (sigmoid);
\draw[connection] (sigmoid) -- (output);
\node [above=0.35cm of sampling, xshift=-1.5cm, yshift=-0.5cm] (zi) {\( z_i \)};

\node[left=0.1cm of sampling, xshift=-0.5cm] (mu) {\( \mu \)};

\node[below=0.3cm of sampling, xshift=0.5cm] (logvar) {\( \log(\text{$\sigma_i^2$}) \)};
\draw[connection] (sampling) -- (logvar);

\end{tikzpicture}

\caption{VAE-LSTM: Variational Autoencoder with LSTM Architecture}
    \label{fig:enter-label}
\end{figure}
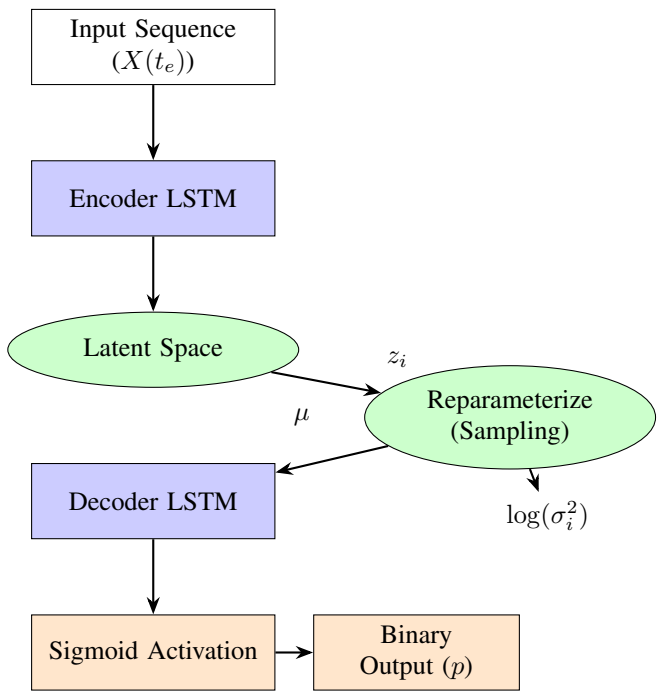

The VAE-LSTM architecture consists of a unidirectional LSTM-based encoder with a single layer and a hidden size of 10 units. The encoder processes input sequences, extracting meaningful representations that are then mapped to a latent space of dimensions $n=10$. This latent space is characterized by the mean $\mu$ and log-variance $\log(\sigma^2)$ of a multivariate Gaussian distribution. The model uses reparameterization to sample from this distribution, introducing stochasticity during training. Then, a unidirectional LSTM-based decoder (a single layer and a hidden size of 10 units) reconstructs the input sequence from the latent representation. The overall architecture is designed to balance the reconstruction loss, computed using binary cross-entropy (BCE), with the KL divergence loss, promoting the learning of meaningful latent representations. 

As given by \eqref{eq:total_loss}, the loss function for this architecture consists of the reconstruction loss $\mathcal{L}_{\text{recon}}$, computed as the BCE loss, and the KL-Divergence loss $\mathcal{L}_{\text{KL}}$.
The BCE loss is appropriate given our binary classification problem, while the KL-Divergence loss regularizes the latent space. 
For BCE computation, the target variable ($y_i$) represents the ground truth labels (0 or 1), $N$ is the number of samples (in batch), and the predicted output ($p_i$) is generated by our VAE-LSTM model. 
The KL-Divergence loss, $\mathcal{L}_{KL}$, is computed using the mean (\(\mu_i\)) and standard deviation (\(\sigma_i\)) of the latent variable $z_i$ for each latent dimension \(n\), for each input sequence.

\begin{equation}
\begin{aligned}
\mathcal{L}_{recon} (BCE) &= -\frac{1}{N} \sum_{i=1}^N [y_i \log(p_i) + (1-y_i) \log(1-p_i)] 
\\
\mathcal{L}_{KL} &= -\frac{1}{2N} \sum_{j=1}^{N} \sum_{i=1}^{n} \left(1 + \log(\sigma_{j,i}^2) - \mu_{j,i}^2 - \sigma_{j,i}^2\right)
\\
\mathcal{L}_{total}&=\mathcal{L}_{recon}+\mathcal{L}_{KL}
\end{aligned}
\label{eq:total_loss}
\end{equation}

\begin{figure*}[t]
      \centering
        \includegraphics[width=\columnwidth,trim={0cm 0.3cm 0cm 0.3cm},clip]{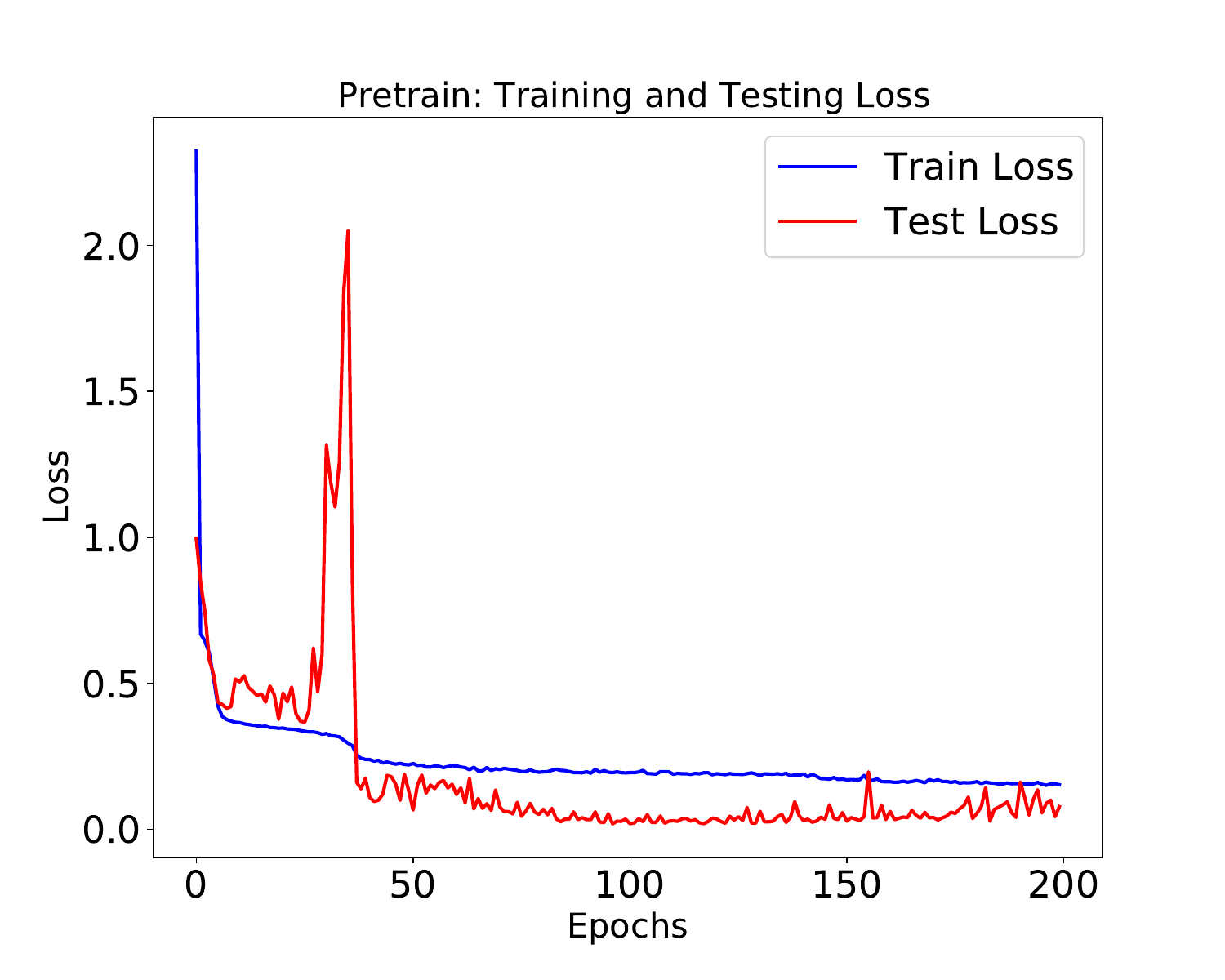}
        \includegraphics[width=\columnwidth,trim={0cm 0.3cm 0.3cm 0.3cm},clip]{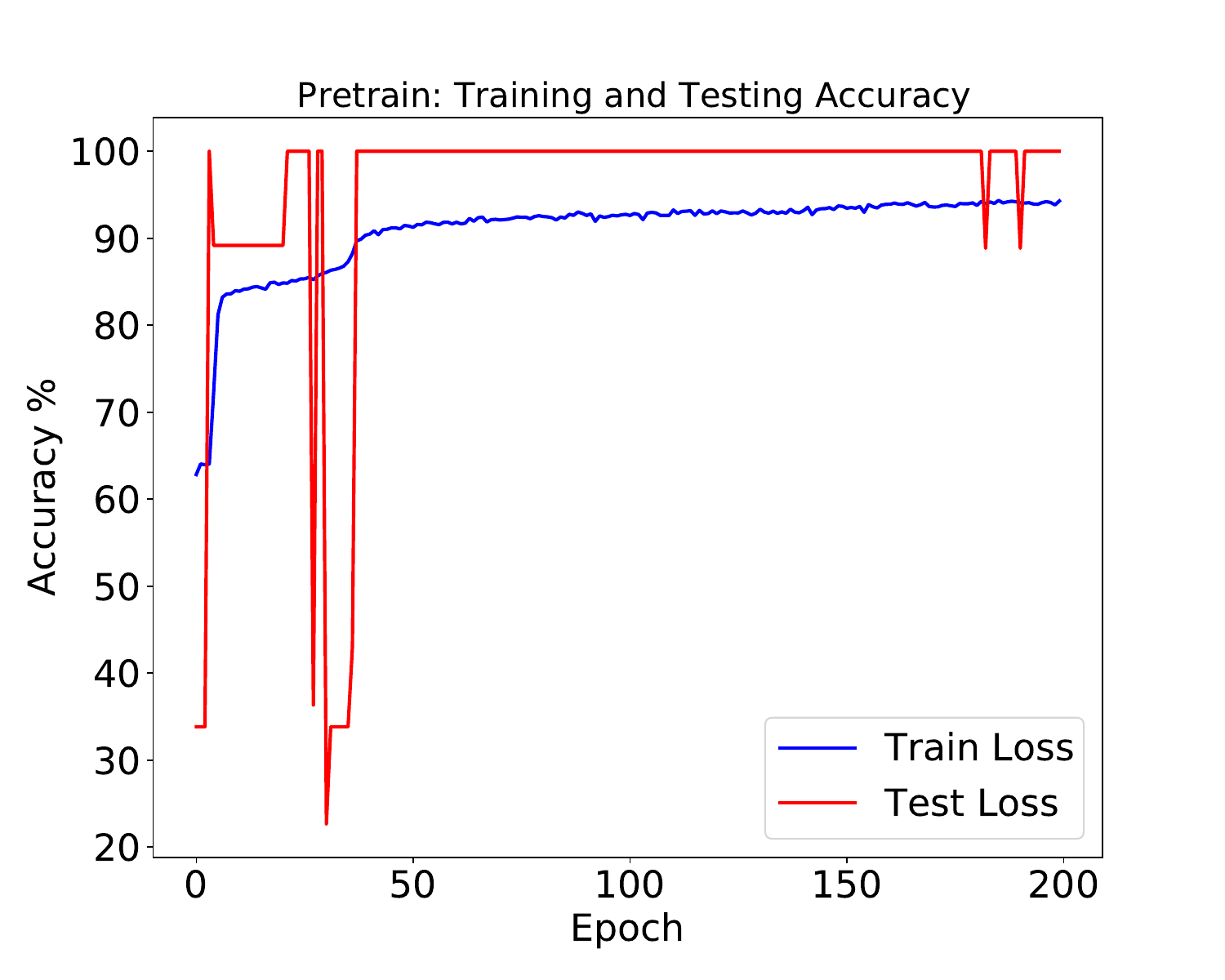}
        \vspace{1cm}
        \includegraphics[width=\columnwidth,trim={0cm 0.3cm 0.3cm 0.3cm},clip]{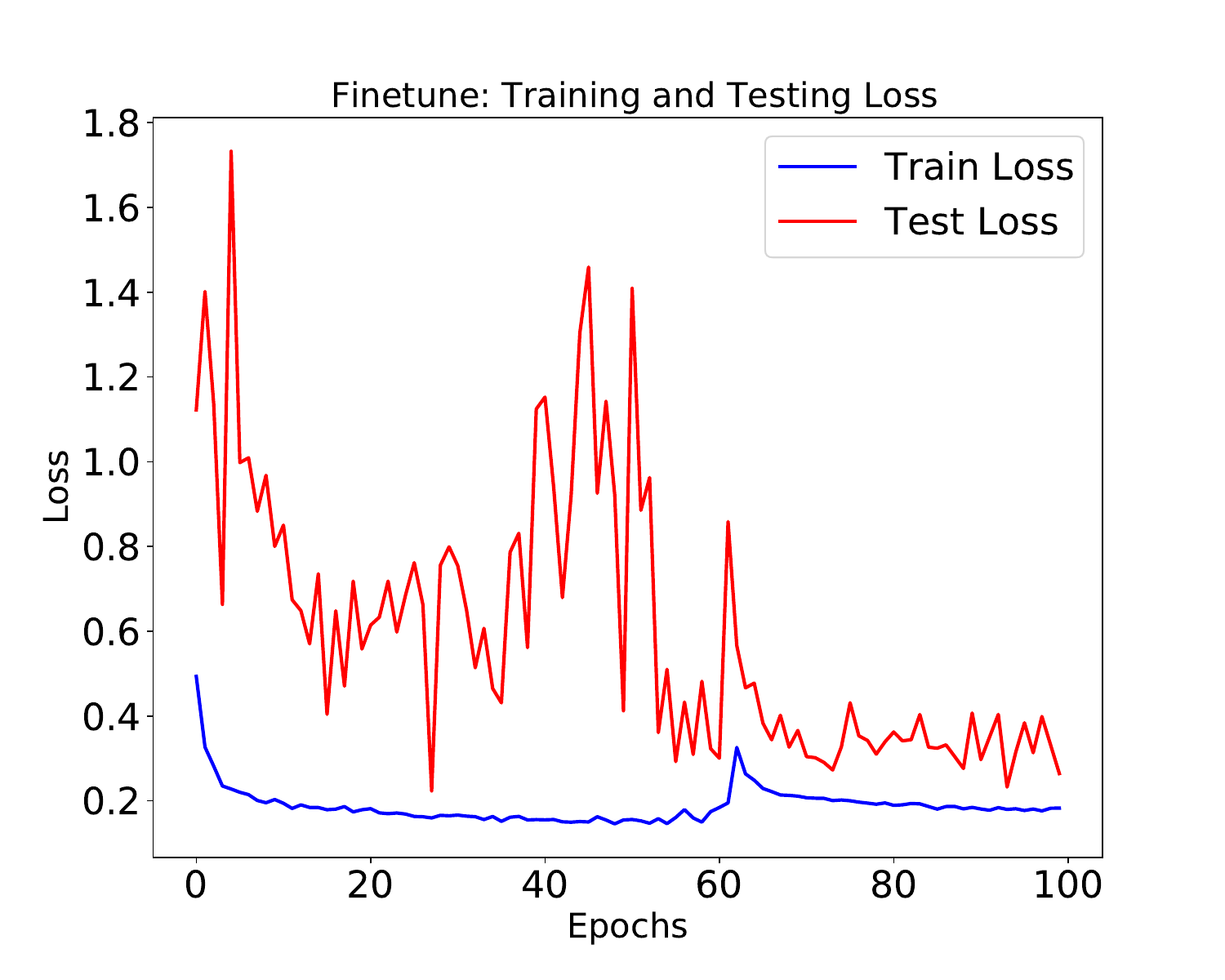}
        \includegraphics[width=\columnwidth,trim={0cm 0.3cm 0.3cm 0.3cm},clip]{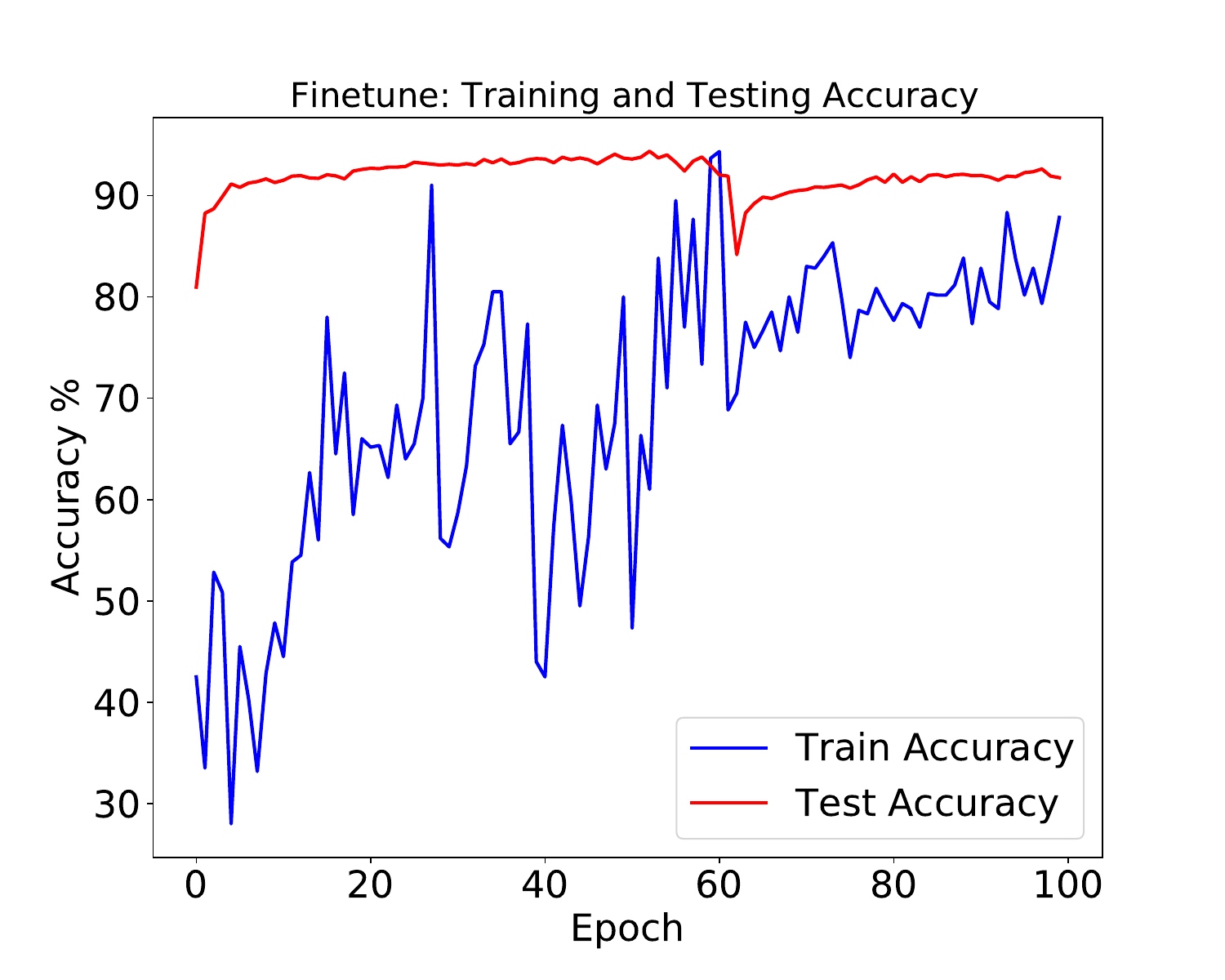}
        
    \setlength\abovecaptionskip{-2.4\baselineskip}
      \caption{LSTM Training Plots, Pretraining: (a) Loss, (b) Accuracy. Finetuning: (a) Loss, (b) Accuracy.}
      \label{fig:AccLoss}
\end{figure*}

The VAE-LSTM architecture employs a two-stage training process to enhance its adaptability to various object weights. In the \textbf{pre-training} phase, the model is initially trained using handovers from the previous dataset, focusing on two weight classes: 0.8 kg and 1.8 kg (Handovers@RPL). Subsequently, during \textbf{the fine-tuning phase}, the pre-trained model is further refined using handovers from the new dataset (Handovers@RPL-2.0), which includes five weight classes: 1.0 kg, 1.2 kg, 1.4 kg, 1.6 kg, and 2.0 kg. We employ a 90/10 train-test split for both the pre-training and fine-tuning phases, a batch size of 100, a learning rate of 0.01 and training for 100 epochs with early stopping based on loss. The model's performance, including accuracy and loss metrics for both training and testing sets, is illustrated in Fig. \ref{fig:AccLoss}.
This two-stage approach enables the trained LSTM to adapt for objects weighing between 0.8 kg and 2.0 kg. By leveraging the reparameterization trick during training, the model introduces stochasticity, allowing it to capture and generate diverse representations of input sequences in a variational and sequential manner. The loss function, combining binary cross-entropy (BCE) for reconstruction and KL divergence, promotes the learning of meaningful latent representations across the range of object weights.

\subsection{Adaptive Robotic Grip Release Strategy}
In a robot-to-human handover, for a robotic giver, the interaction forces and the weight of the object in the handover can be measured at a wrist sensor to measure the following time series:
 \begin{equation}
 X^R(t_e)= \{F_{y}^{t}, F_{z}^{t}, w  : t \in [t_e\text{-} 100,t_e]\}.
 \label{eqnR:time_series_Int_wrench}
 \end{equation} 
$X^R(t_e)$ is then input to the trained VAE-LSTM to command a human-inspired grip-release for object weights 0.8 kg to 1.8 kg. For lighter objects weighing 0.0 to 0.8 kg, which are currently not in the training dataset for VAE-LSTM, we employ a pull force-based threshold ($F_{y}<4 N$), which was shown to be most effective for lightweight objects in our previous study \cite{parag-humanoids}. This hybrid approach allows for adaptive grip release across a wider range of object weights, addressing the limitations of a single strategy and potentially improving both efficiency and perceived naturalness of robot-human handovers. We verified that a forward pass of this VAE-LSTM model implemented via a PyTorch-CUDA framework running on an Nvidia GeForce RTX 3070 Super Max-Q GPU took less than the 8.333 ms needed for the system frequency of 120 Hz.

\subsection{Robot to Human handovers experimentation}
In our human-robot handover experiments, we aimed to evaluate the proposed adaptive grip release strategy during robot-to-human transfers. We utilized the aforementioned identical-looking boxes, each containing a different weight, to ensure participants could not visually discern the weight variations. The boxes weighed 90 g for the Low, 0.9 kg for the Moderate and 1180g for the Heavy category.

\subsubsection{Baselines}
We will compare our adaptive strategy against two baseline strategies:
\begin{itemize}

    \item Load share-based grip release, where grip release occurs when load share falls below 50\%.
    \item Pull force-based grip release, defined by a threshold of $F_{y} < 4 \, \text{N}$.
\end{itemize}

\begin{figure}[h!]
      \centering
       \includegraphics[width=0.48\linewidth,trim={3.2cm 8.5cm 4.2cm 8.3cm},clip]{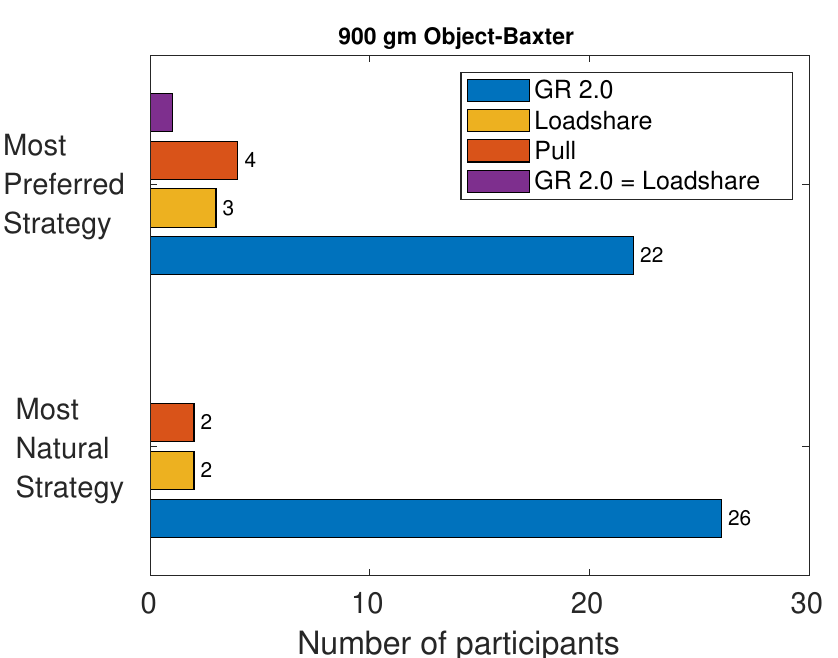}  \includegraphics[width=0.48\linewidth,trim={3.2cm 8.5cm 4.2cm 8.3cm},clip]{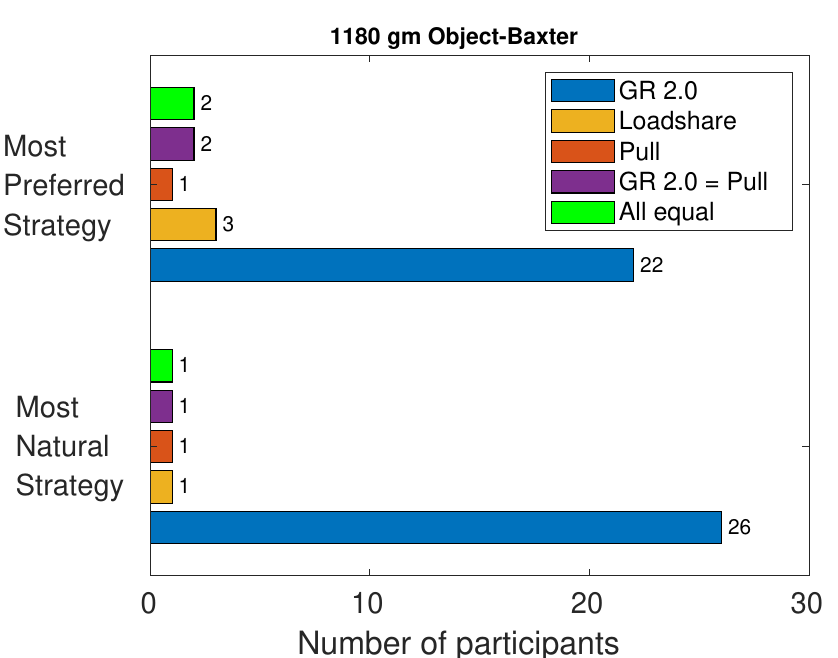}
       \includegraphics[width=0.48\linewidth,trim={3.2cm 8.5cm 4.2cm 8.3cm},clip]{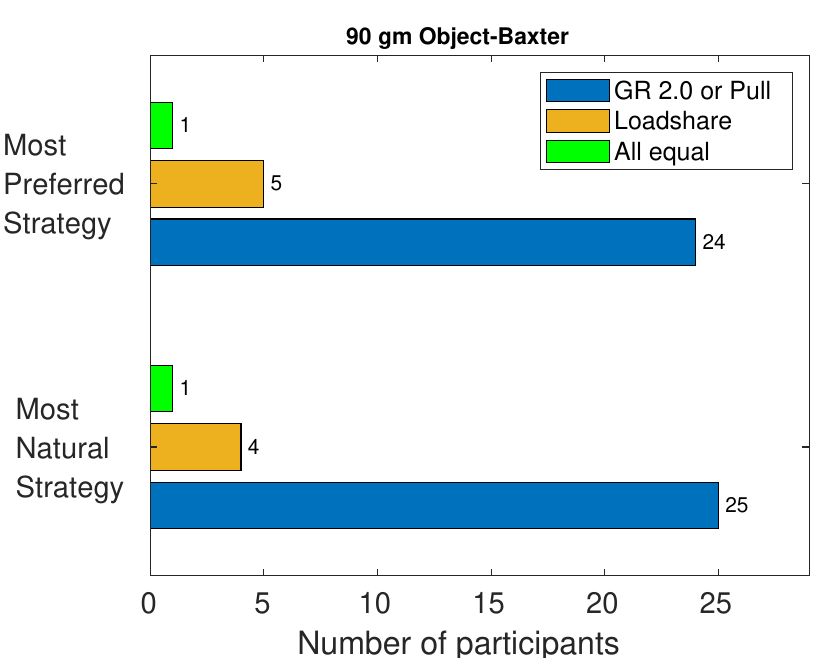}
    \caption{Responses comparing strategies for different Objects}
      \label{figT:Answers_obj_800_Baxter}
   \end{figure}
\subsubsection{Method}
We test our proposed data-driven adaptive grip release strategy that adjusts according to the object weight during robot-to-human handovers.
In the robot-to-human handover experimentation, the grip-release strategies were labeled as:
\begin{itemize}
    \item Strategy A: Grip Release Strategy 2.0 (GR 2.0)
    \item Strategy B: Load-Share-Based Strategy
    \item Strategy C: Pull Force based Strategy
\end{itemize}
The objects used in the experiment were labeled as Object 1, Object 2, and Object 3, weighing 0.09 kg, 0.9  kg, and 1.18 kg, respectively.
In robot-to-human handovers, participants took the objects from the Baxter robot once it reached the handover phase. 
The robotic motion leading up to the object handover was identical for each object in each grip-release strategy, correspinding to the baseline motion in Section \ref{section:adaptive_motion}.B. 
No other details about the strategies were provided to the participants, except naming them as A, B, and C. 
Each participant performed these handovers one object at a time encountering the three strategies for each object.
They were was asked to take each object from the robot four times for each of the strategies in sequence, to ensure consistency and reliability in the results. 
For subjective analysis, participants answered a 7-point Likert scale questionnaire after the handovers with each strategy. In particular, for each strategy the questions evaluted differnt parameters:
\begin{enumerate}
    \item How easy it was to take the object during the handover (1: Very Hard, 7: Very Easy).
\item How likely the object was to drop during the handover (1: Likely to drop, 7: Not likely to drop).
\item How natural the handover felt (1: Not natural at all, 7: Very natural).
\item How safe the handover felt (1: Not safe at all, 7: Very safe).
\end{enumerate}
After experiencing all three strategies for a particular object, participants answered additional questions to compare the stratgies and choose:
\begin{enumerate}
\item Which strategy felt the most natural (A, B, or C)?
\item Which strategy they preferred (A, B, or C)?
\item Which strategy felt the safest (A, B, or C)?
\end{enumerate}
For quantitative analysis, the strategies were evaluated based on how early they triggered the grip release during the handover phase.

\textbf{Hypotheses:} We hypothesize the following for robot-to-human handovers:  
\begin{itemize}
    \item \textbf{HG1:} The human-inspired strategy GR 2.0 results in more natural handovers compared to the baselines.
    \item \textbf{HG2:} Handovers using the GR 2.0 strategy are perceived as easier for the receiver to take the object.
    \item \textbf{HG3:} The GR 2.0 strategy reduces the likelihood of object dropping during handovers.
    \item \textbf{HG4:} Handovers with the GR 2.0 strategy are perceived as safer.
\end{itemize}
\subsubsection{Results}
After a participant completed the experimentation testing the adaptive robotic motion, they procedded to do the experimentation for the Grip Release strategies. 
For all recorded handovers, even when a specific strategy was being tested for commanding grip release, we also documented and assessed when and if other strategies would have triggered a grip release.

\begin{figure}[h!]
\centering
\subfloat[For Pull force-based strategy]{
\includegraphics[width=.75\linewidth,trim={2.2cm 7.3cm 3.0cm 8.3cm},clip]{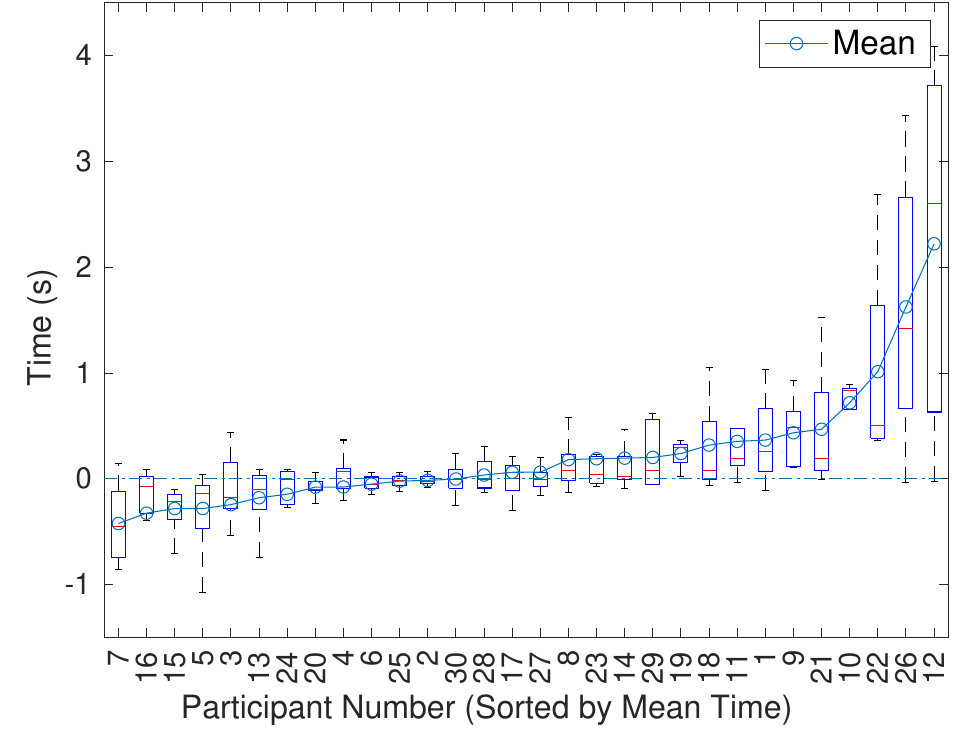}}
\vspace{1mm}
\subfloat[For Loadshare-based strategy]{
\includegraphics[width=.75\linewidth,trim={2.2cm 7.3cm 3.0cm 8.2cm},clip]{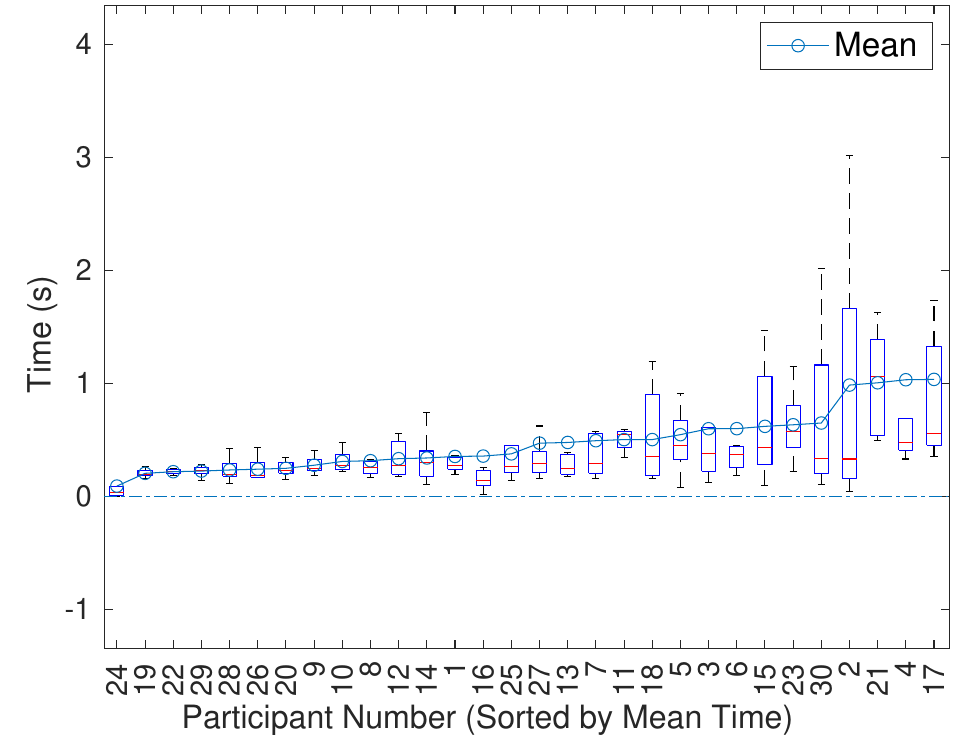}}
\setlength\abovecaptionskip{0.15\baselineskip}
\caption{Object 0.9 kg: Difference in time of grip release command relative to GR 2.0 strategy.}
\label{figC:new_BCto_A_obj1}
\end{figure}

\begin{figure}[h!]
\centering
\subfloat[For Pull force-based strategy]{
\includegraphics[width=.75\linewidth,trim={2.2cm 7.3cm 1.6cm 8.3cm},clip]{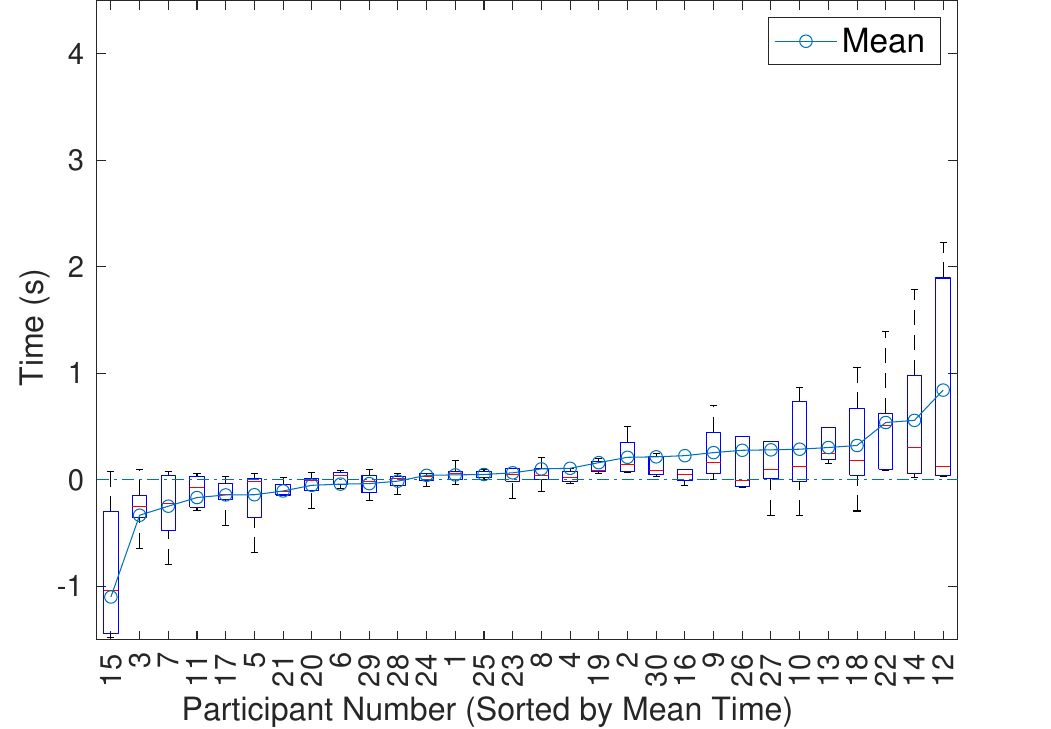}}
\vspace{1mm}
\subfloat[For Loadshare-based strategy]{
\includegraphics[width=.75\linewidth,trim={2.2cm 7.3cm 1.6cm 8.1cm},clip]{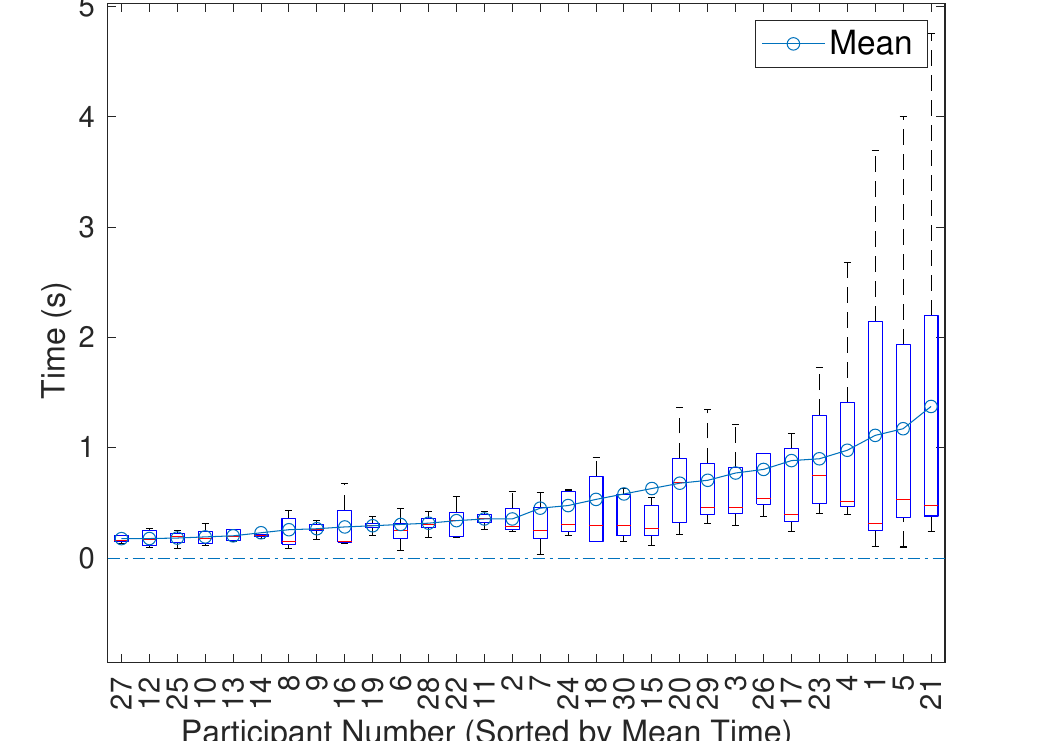}}
\setlength\abovecaptionskip{0.15\baselineskip}
\caption{Object 1.180 kg: Difference in time of grip release command relative to GR 2.0 strategy.}
\label{figC:new_BCto_A_obj2}
\end{figure}

The majority of participants preferred the GR 2.0 strategy for moderate and heavy objects (0.9 kg and 1.18 kg), finding it the most natural, as illustrated in Fig.~\ref{figC:Answers_obj_800_Baxter}. This preference was further reinforced by the observation that GR 2.0 allowed for quicker grip release compared to the other two strategies, as shown in Fig.~\ref{figC:new_BCto_A_obj1} and Fig.~\ref{figC:new_BCto_A_obj2}. An analysis of the average differences in grip release timing commands revealed that GR 2.0 consistently resulted in faster releases for all participants relative to the loadshare-based release method. Furthermore, GR 2.0 facilitated quicker releases for most participants compared to the pull force-based grip release strategy.
This analysis also suggests that the pull force-based release outperformed the loadshare-based release. 
Overall, for the moderate object, GR 2.0 was the fastest for 225 (62.3\%), while the pull force-based release was the fastest for 135 (37.4\%), and the loadshare-based release was the fastest for only 1 of the recorded handovers. For the heavy object, GR 2.0 was the fastest for 255 (70.83\%), whereas the pull force-based release was the fastest for the remaining 105 (29.17\%) of the recorded handovers.

Interestingly, for the moderate-weight object, the pull force-based release was not triggered in 92 instances, the loadshare-based release in 10 instances, and GR 2.0 in only 1 instance. Similarly, for the heavy object, the pull force-based release was not triggered in 85 instances, the loadshare-based release in 16 instances, and GR 2.0 in just 2 instances.  

These results further highlight the effectiveness of the GR 2.0 strategy. They also suggest that although the loadshare-based release tends to be slower, it is more reliable than the pull force-based strategy in detecting the taker's interaction during a handover and commanding grip release. The fact that pull force was not triggered—indicating the pull force threshold was not met during those handovers, as grip release was commanded by another strategy—suggests that relying solely on pull force may not be a dependable approach, as it does not consistently capture the taker's intent to take the object.


For the light object (0.09 kg or 90g), participant preferences were mixed between Grip Release Strategy 2.0 and the Pull Force-Based Strategy, both of which rely on the same pull force threshold for lighter objects. However, false grip releases were sometimes caused by the loadshare-based strategy for the light object. This occurred because, due to the object's low weight, the loadshare forces remained below the interaction sensor's noise threshold. As a result, the object was sometimes released before participants reached it. Although no objects were dropped, participants perceived the robot as ``handing over the object in its gripper without actually gripping it.''

\begin{figure}[h]
\centering
\includegraphics[width=.75\linewidth,trim={1.8cm 7.3cm 2.6cm 8.2cm},clip]{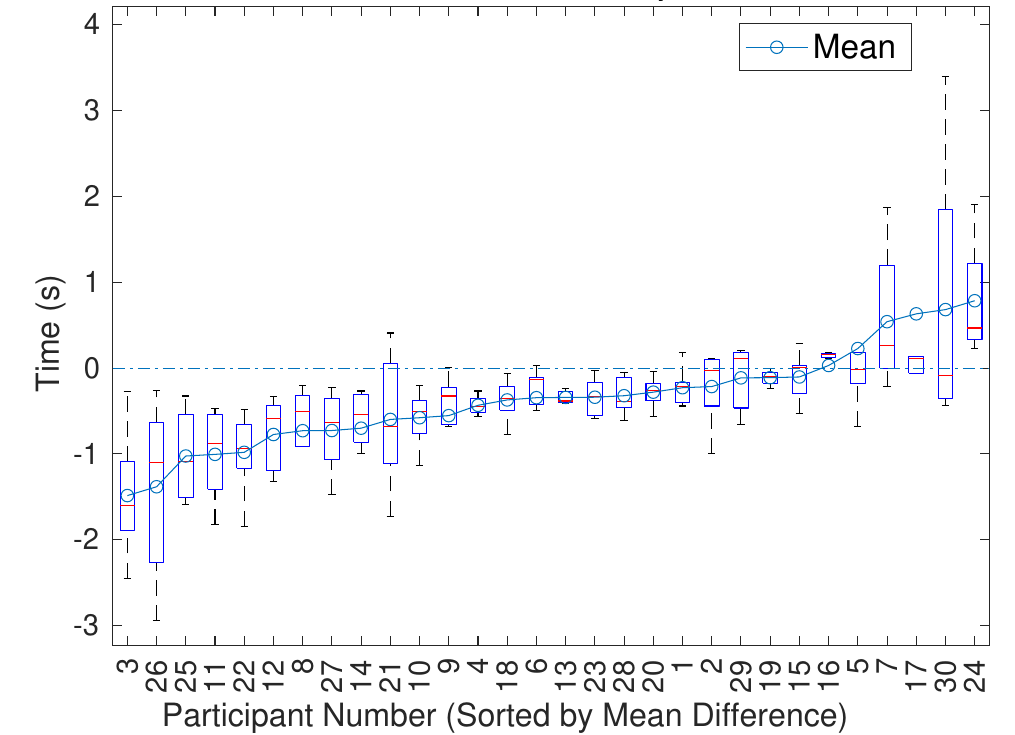}
\caption{Object: 0.09 kg: Difference in time of grip release command for the loadshare-based strategy relative to the pull force-Based strategy.}
\label{figC:new_BCto_A_obj3}
\end{figure}
When false grip releases did not occur, the loadshare-based strategy led to very fast grip releases, almost as soon as the participant touched the object. As seen in Fig.~\ref{figC:new_BCto_A_obj3}, the loadshare-based strategy led to faster grip release commands than the pull force-based strategy for most participants. This resulted in an increase in ease of taking the object but a decrease in perceived safety. Several participants reported that this release was too fast, with almost no resistance, which did not feel human-like. Consequently, most participants perceived a higher likelihood of object drops compared to the other strategies. As a result, the loadshare-based strategy was not considered the most natural or preferred by most participants. However, some participants liked the very quick release behavior due to the ease of taking the object.

Interestingly, when the loadshare-based strategy was used to command the handovers, the pull force-based release was not triggered for 90 handovers out of 112, implying that the pull force threshold was seldom reached when the loadshare-based grip release occurred for the light object. The loadshare-based grip release did not trigger for 18 handovers out of 226 total handovers when the pull force-based strategy was active.
\textbf{ANOVA and Pairwise t-tests for the three Handover Strategies}:

We conducted a one-way ANOVA to compare the 3 strategies A (GR 2.0), B (Loadshare) and C (Pull) on the answers of participants for the questionnaire on ease of taking, naturalness, likeliness to drop and perceived safety. This was followed by post-hoc pairwise t-tests for further analysis.

\subsubsection*{Object 1-0.9kg}
\begin{figure}[h!]
      \centering
       \includegraphics[width=0.9\linewidth,trim={1.2cm 6.3cm 1.7cm 5.8cm},clip]{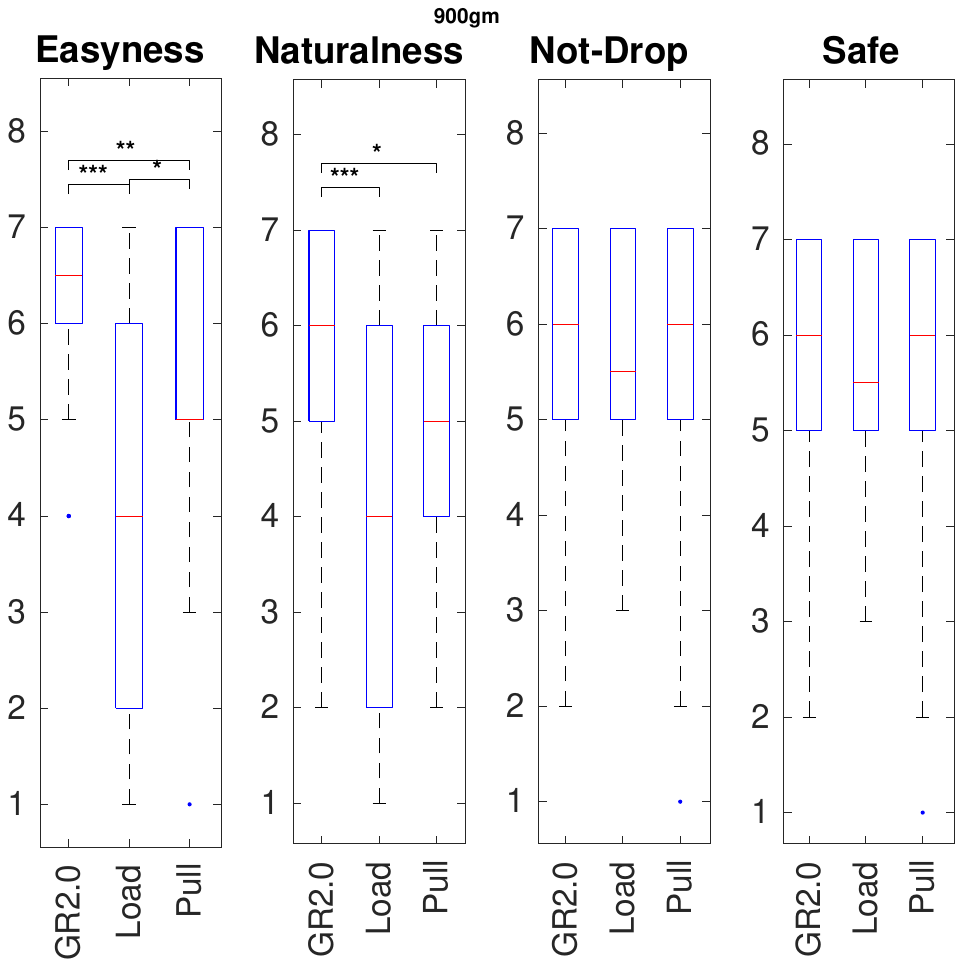}
    \caption{Responses comparing strategies Object-1 0.9 kg}
      \label{figC:Answers_obj_900_Baxter}
   \end{figure}
\begin{itemize}
    \item \textbf{How easy was it to take the object in handover:} 
    ANOVA shows a significant difference ($p < 0.001$), with pairwise t-tests indicating that all comparisons between strategies are significant, with GR 2.0 being the highest scored and Loadshare-based being the lowest scored. 

    \item \textbf{How natural the handover felt:} 
    ANOVA is significant ($p = 0.002$), with a significant difference found between GR 2.0 strategy and Loadshare-based strategy as well as GR 2.0 and Pull-based strategies in pairwise t-tests. The GR 2.0 was the highest scored for naturalness.
    
    \item \textbf{How likely was that the object would drop during handover:} 
    No significant difference found in either ANOVA ($p = 0.965$) or pairwise t-tests.
    
    \item \textbf{How safe did the handover feel:} 
    ANOVA results show no significant difference ($p = 0.111$), with no significant differences in the pairwise t-tests.
\end{itemize}
These results support HG1 and HG2 for object 1.

\begin{figure}[h!]
      \centering
       \includegraphics[width=0.9\linewidth,trim={1.2cm 6.3cm 1.7cm 5.8cm},clip]{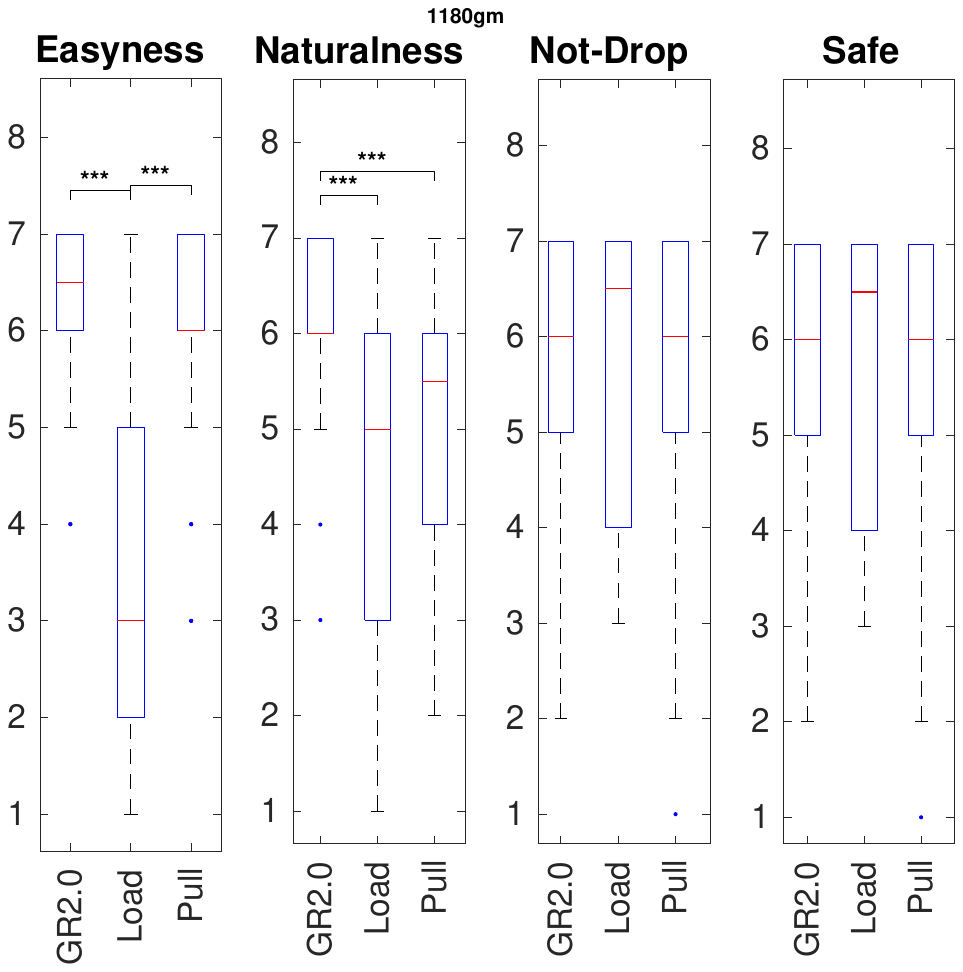}
    \caption{Responses comparing strategies Object-2 1.180 kg}
      \label{figC:Answers_obj_1180_Baxter}
   \end{figure}

\subsubsection*{Object 2- 1.180 kg (Fig. \ref{figC:Answers_obj_1180_Baxter})}

\begin{itemize}
    \item \textbf{How easy was it to take the object in handover:} 
    ANOVA shows a significant difference ($p < 0.001$), with significant pairwise differences between GR 2.0 strategy and Loadshare-based strategy, as well as Loadshare-based and Pull-based strategies. No significant difference between GR 2.0 and Pull-based strategies. The loadshare-based strategy was scored the lowest, while GR 2.0 was scored a little higher than pull force-based strategy.
    
    \item \textbf{How natural the handover felt:} 
    ANOVA is significant ($p < 0.001$), with significant pairwise differences between GR 2.0 vs Loadshare- based strategies and GR 2.0 vs Pull-based strategies, but no significant difference between Loadshare and Pull-based strategies. GR 2.0 was scored the highest among the three. 

    \item \textbf{How likely was that the object would drop during handover:} 
    No significant differences found in ANOVA ($p = 0.558$) or pairwise t-tests.
    
    \item \textbf{How safe did the handover feel:} 
    No significant differences in either ANOVA ($p = 0.304$) or pairwise t-tests.
\end{itemize}
These results support HG1 for object 2.

\begin{figure}[h!]
      \centering
       \includegraphics[width=0.9\linewidth,trim={1.2cm 6.7cm 1.7cm 6.3cm},clip]{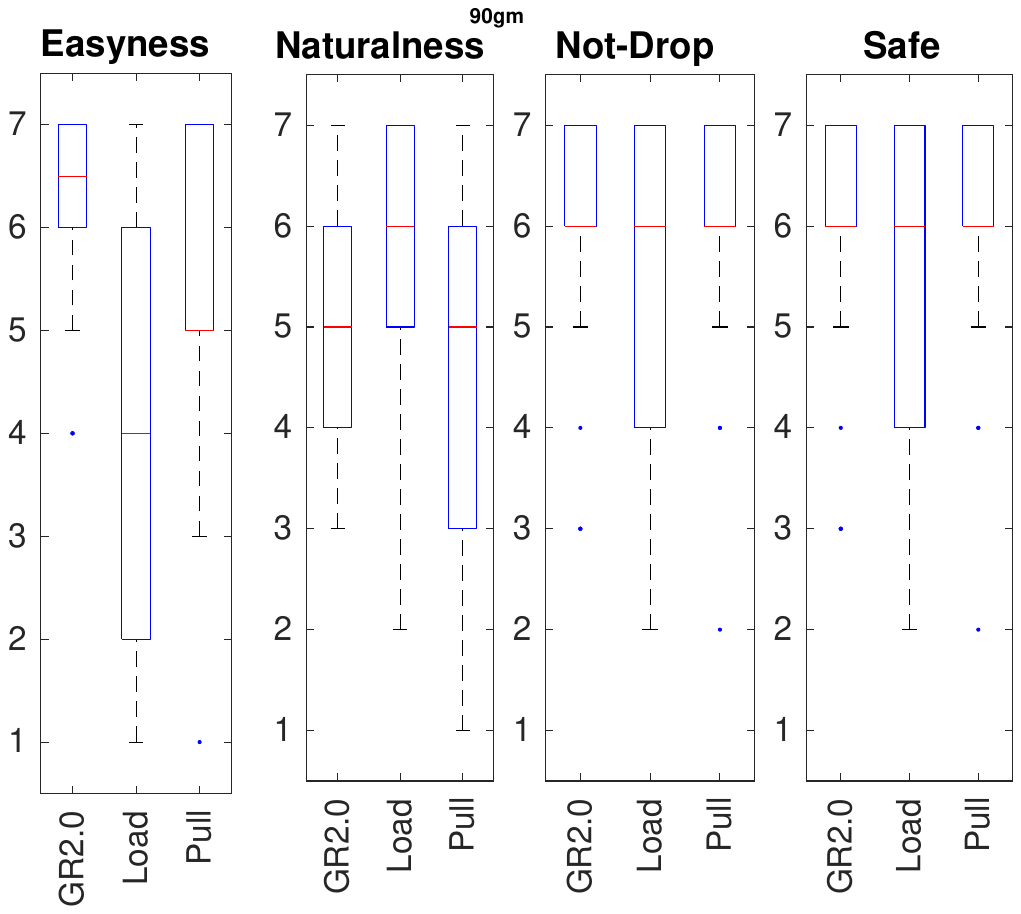}
    \caption{Responses comparing strategies Object-3 0.090 kg}
      \label{figC:Answers_obj_90_Baxter}
   \end{figure}

\subsubsection*{Object 3 - 0.09 kg (Fig. \ref{figC:Answers_obj_90_Baxter})}
\begin{itemize}
    \item \textbf{How easy was it to take the object in handover:} 
    ANOVA shows no significant difference ($p = 0.216$), with no significant pairwise differences.
    
    \item \textbf{How likely was that the object would drop during handover:} 
    No significant differences found in ANOVA ($p = 0.167$) or pairwise t-tests.
    
    \item \textbf{How natural the handover felt:} 
    ANOVA results show no significant difference ($p = 0.133$), and pairwise t-tests confirm no significant differences.
    
    \item \textbf{How safe did the handover feel:} 
    ANOVA results are not significant ($p = 0.966$), with no significant pairwise differences.
\end{itemize}


The analysis of handover performance across three objects reveals important insights into the effectiveness of different strategies. For the heavier objects of 0.9 and 1.180 kg, the GR 2.0 (Strategy A) generally outperformed the other strategies, particularly in the ease of taking the object and the naturalness of the handover, with significant differences observed between the GR 2.0 (A) and loadshare-based strategy (B). 
Loadshare-based Strategy B consistently received the lowest ratings, especially for the ease of taking the object and the naturalness of the handover. Pull Force-based grip release strategy C, performed in between GR 2.0 and Loadshare-based Strategy, showing mixed results. 
Thus, we found \textbf{HG1} and \textbf{HG2} to be valid for Object 1, and \textbf{HG1} to be valid for Object 2, suggesting that GR 2.0 leads to more natural and easier handovers.
However, for Object 3 (90 g), the lighter weight led to no significant differences between the strategies in any of the measured categories, indicating that the strategy used may have less impact on lighter objects. 

Furthermore, our results do not support \textbf{HG3} or \textbf{HG4} for any objects, indicating no effect of different grip-release strategies on perceived safety or the likelihood of object dropping. These findings may be attributed to the inherently safe design of the handovers in our experiments, where the object was prevented from falling out of the robot's gripper.

Overall, GR 2.0 appears to be the most effective strategy for handling heavier objects, with loadshare-based strategies generally underperforming in most cases.

\section{Conclusion and Future Work}

In this work, we explored the role of object weight in influencing human motion and grip-release strategies during handovers, using these insights to inspire strategies for robotic systems. 
To analyze human handovers, we recorded different datasets: \textbf{Handovers@RPL}, which included datasets with a special object to record forces and motion together in handovers, and \textbf{YCB-Handovers}, which included handovers with generic objects from the YCB dataset. 

We found that object weight significantly affects key parameters such as transfer time, grip release time, and interaction forces.
Moreover, object weight influenced human hand motion in human-human handovers, showing a moderate negative correlation between weight and motion, i.e., an increase in object weight leads to a decrease in velocity and acceleration.
Additionally, we grouped objects into three categories—\textit{Low, Moderate, and High}—based on observed differences in velocities and accelerations.

Furthermore, we developed adaptive robotic strategies for motion and grip-release, aiming to improve the naturalness, efficiency, and safety of robot-to-human handovers. 
Our proposed strategies, inspired by human-human interactions, were evaluated through a series of user studies with objects belonging to the three weight categories. 
The adaptive motion strategy, which reduced maximum acceleration similar to humans with increasing weight, allowed participants to better perceive the object's weight during robot-to-human handovers. 
Additionally, the adaptive grip-release strategy, a data-driven approach based on human grip release patterns, demonstrated improved performance for medium and heavy-weight objects compared to the baselines.

Despite these promising results, there are some limitations to this work. First, the adaptive motion strategy relied on predefined maximum and average acceleration values, which were constrained by the safety limits of the robot. While found effective, these fixed parameters may not fully capture the nuanced dynamics of human-human handovers. Thus, the future work could focus on using robots that could replicate the exact motion parameters observed for the three weight classes in human handovers and learning-based approaches that adaptively refine these parameters in real-time based on participant feedback or environmental context. Second, the adaptive grip-release strategy faced challenges with lightweight objects, where due to lack of data, a data driven strategy could not be formulated for the objects below 0.8 kg weight. This can be addressed by using different light weight sensors to collect the force data in human handovers with lower object weights.
Further, incorporating additional modalities, such as vision or tactile feedback, could enhance the robustness of the grip-release strategy.

In addition, while our datasets provide valuable insights into weight-based adaptations, they are limited to specific handover scenario and experimental setups. Expanding the datasets to include a broader range of handover scenarios, and participant demographics will be helpful to ensure the generalizability of our findings. Finally, investigating the integration of adaptive motion and grip-release strategies into a cohesive framework could further advance the state of human-robot handover systems, paving the way for more intuitive and effective human-robot collaboration in real-world scenarios.

\section*{ACKNOWLEDGMENT}
This work was supported by Digital Futures at KTH. 
\vspace{0.1mm}
\vspace{-2mm}
\bibliographystyle{IEEEtran}
\bibliography{IEEEabrv,pk_ref_paper_4_handovers.bib}
\end{document}